\documentclass[onefignum,onetabnum]{siamonline190516}
\usepackage{amssymb,amsmath}
\usepackage[utf8]{inputenc}
\usepackage{relsize,mathtools,nccmath} 

\usepackage{lineno}

\usepackage{algorithmic}
\usepackage{multirow}

\newcommand{\edits}[1]{\textcolor{black}{#1}}

\newcommand{\hf}{{\frac 12}}

\newcommand{\arras}{{\stackrel{a.s.}{\longrightarrow}}}

\newcommand{\grad}{\ensuremath{\nabla}}

\newcommand{\bfC}{{\bf C}}

\newcommand{\bfH}{{\bf H}}
\newcommand{\bfI}{{\bf I}}

\newcommand{\bfK}{{\bf K}}

\newcommand{\bfP}{{\bf P}}
\newcommand{\bfQ}{{\bf Q}}

\newcommand{\bfS}{{\bf S}}

\newcommand{\bfX}{{\bf X}}

\newcommand{\bfZ}{{\bf Z}}

\newcommand{\bfb}{{\bf b}}

\newcommand{\bfg}{{\bf g}}

\newcommand{\bfx}{{\bf x}}

\newcommand{\bfu}{{\bf u}}
\newcommand{\bfq}{{\bf q}}

\newcommand{\bfd}{{\bf d}}

\newcommand{\bfr}{{\bf r}}

\newcommand{\bfw}{{\bf w}}
\newcommand{\bfz}{{\bf z}}

\newcommand{\bfepsilon}{{\boldsymbol \varepsilon}}
\newcommand{\bftheta}{{\boldsymbol \theta}}
\newcommand{\bfTheta}{{\boldsymbol \Theta}}

\newcommand{\bfomega}{{\boldsymbol \omega}}

\usepackage{tikz}
\usepackage{amsfonts}
\usepackage{graphicx}

\usepackage{enumitem}
\setlist[enumerate]{leftmargin=.5in}
\setlist[itemize]{leftmargin=.5in}

\newsiamremark{example}{Example}

\title{Estimating a potential without the agony of the partition function}

\author{Eldad Haber, Moshe Eliasof and Luis Tenorio}

\begin{document}

\maketitle

\begin{abstract}
Estimating a Gibbs density function given a sample is an
important problem in computational statistics and statistical learning. Although
the well established  maximum likelihood method is commonly used, it requires the computation of the partition function (i.e., the normalization of the density).
 This function can be easily calculated for simple low-dimensional problems but its computation is difficult or even intractable for general densities
and high-dimensional problems.
In this paper we propose an alternative approach based on Maximum A-Posteriori (MAP) estimators, we name Maximum Recovery MAP (MR-MAP),  to derive  estimators that do not require the computation of the partition function, and reformulate the problem  as an optimization problem. We further propose a least-action type potential that allows us to quickly solve the optimization problem as a feed-forward hyperbolic neural network.
We demonstrate the effectiveness of our methods on some standard data sets.
 
\end{abstract}



\section{Introduction}\label{sec1}
We consider the problem of estimating a probability density function $p_\bftheta$ on $\mathbb{R}^p$ given independent and identically distributed
(i.i.d) samples $\bfx_1,\ldots,\bfx_n$. The density (with respect to the Lebesgue measure on $\mathbb{R}^p$) is assumed to be of the general parametric form
\begin{equation}\label{eq:dist}
p_{\bftheta}(\bfx) = {\frac 1{Z(\bftheta)}}  e^{-\phi(\bfx,\bftheta)},
\end{equation}
where $\phi$ is often called the potential function and $Z$ the partition function.
This type of problem arises naturally in fields such as computational biology \cite{SchaferStrimmer05}, statistical physics \cite{crooks2007measuring}, image processing \cite{jinggang}, image synthesis \cite{gao2021learning,zwm}
and more.
In some applications such as Newtonian mechanics the potential and its parameters are known, but most often it is unknown or \edits{its parametric form is known but the parameters $\bftheta$ are unknown}. To estimate the functional form of $\phi$ and its
parameters $\bftheta$, or even to determine $\bftheta$ given a parametric form
of $\phi$, requires dealing with the partition function or its gradient.

 For example, suppose $\phi$ is known up to $\bftheta$ and we use maximum likelihood 
 to estimate $\bftheta$. Let $\bfX = [\,\bfx_1 \mid \cdots   \mid \bfx_n\,]$ be the matrix whose columns are the i.i.d data  with density $p_\bftheta$.
 A maximum likelihood
estimate $\widehat{\bftheta}$ is obtained by maximizing the likelihood
$L(\bftheta;\bfX)= p_{\bftheta}(\bfx_1)\cdots p_{\bftheta}(\bfx_n)$ over $\bftheta$, which is equivalent to minimizing the normalized 
negative log-likelihood $\ell(\bftheta,\bfX) = -\log L(\bftheta;\bfX)/n$ (excluding irrelevant constants):
\begin{eqnarray}
\label{minlog} 
\widehat \bftheta = {\rm arg}\min_{\bftheta}\,\ell(\bftheta,\bfX)
= {\rm arg}\min_{\bftheta}\,
\log Z(\bftheta) + {\frac 1n} \medmath{\sum_{i=1}^n }\, \phi(\bfx_i,\bftheta).
\end{eqnarray}
A major difficulty with computing this estimator stems from the calculation of the partition function and its gradient. The gradient is important as most algorithms use a gradient based optimization (typically stochastic gradient descent \cite{robbins1951stochastic}).
One way to
approximate this gradient is using a sample average of the potential gradient 
that is justified by properties of the score function. Recall that the score function associated to the likelihood based on a sample $\bfx$ with density $p_{\bftheta}$ is defined as
$\bfS(\bftheta,\bfx) = \grad_{\bftheta}\log p_{\bftheta}(\bfx)$. Under the usual regularity conditions used to study Fisher information and properties of  maximum likelihood (see \cite{schervish}[Sec.2.3]) \edits{and assuming $\bfx$ has density $p_{\bftheta}$}, we have
$\mathbb{E}\, \bfS(\bftheta,\bfx)=\boldsymbol{0}$ (the variance is the Fisher information matrix), which yields
\[
\mathbb{E}\,\grad_{\bftheta} \phi(\bfx, \bftheta)= -\grad_{\bftheta}\log Z(\bftheta).
\]
The gradient of  $\ell(\bftheta,\bfX)$ is then given by
\begin{eqnarray}
\label{gradminlog} 
\bfg(\bftheta) =  \grad_{\bftheta}\,\ell(\bftheta;\bfX)
=
-{\mathbb E}\,\grad_{\bftheta} \phi(\bfx, \bftheta)   +{\frac 1n} 
\medmath{\sum_{i=1}^n}\, \grad_{\bftheta} \phi(\bfx_i,\bftheta).
\end{eqnarray}
For any given $\bftheta$, $\bfg(\bftheta)$ can be  approximated using a sample average of the expectation above:
\begin{equation}\label{eq:ex-gradest}
\widehat{\bfg}(\bftheta) = -\frac{1}{N}\medmath{\sum_{j=1}^N}\,\grad_{\bftheta} \phi(\widetilde{\bfx}_j, \bftheta) + {\frac 1n} \medmath{\sum_{i=1}^n }\, \grad_{\bftheta} \phi(\bfx_i,\bftheta),
\end{equation}
where $\widetilde{\bfx}_1,\ldots,\widetilde{\bfx}_N$ are i.i.d with density $p_{\bftheta}$. These samples may be difficult to obtain when the distribution is high-dimensional. Techniques such as Markov chain Monte Carlo (MCMC) and Langevin dynamics are often used but they tend to converge slowly \cite{lin2022on}. Thus, maximum likelihood estimation for large-scale problems is a difficult and sometimes intractable problem.

Another difficulty that is not often discussed is the numerical stability of the method. Consider a potential $\phi$ that is very flat as a function of $\theta$. In this case
$\nabla_\bftheta\phi$ is approximately zero regardless of $\bfx$, causing the method to stall or converge to a trivial
solution. This may not be an issue 
for problems where the dependence on $\bftheta$ is not nearly constant over a ``reasonable" domain of $\bftheta$. But when $\phi$ is a neural network with millions of degrees of freedom, obtaining a solution $\phi(\bfx, \bftheta)$ that is approximately constant in $\bftheta$ 
over a large domain can certainly occur, resulting in an unstable method.

These difficulties are in the core of the work
(see \cite{desjardins2011tracking,gao2021learning,KRAUSE2020103195,lin2022on} and references therein), where different techniques have been proposed to estimate the partition function. In particular, 
in \cite{desjardins2011tracking} a way to `track' the computation of the partition function has been proposed, in \cite{lin2022on} the convergence of the Monte Carlo
integration  was studied, and in \cite{gao2021learning} a method based on properties of a joint distribution with latent data is used to speed  up the computations. Nonetheless, all methods known to us need to address the computation of the partition function (or its gradient) and this is difficult to do when the distribution is highly non-local or multi-modal.  
We present a simple example to  demonstrate some of these difficulties 

\begin{example}\label{ex:precision-mat}
To illustrate  the computational problems we may face in finding maximum likelihood estimates, we turn to the problem of covariance estimation where the partition function and its gradient can be computed analytically. 
Assume that  $p_{\bfTheta}$ is the density of $N_p(\boldsymbol{0},\bfTheta^{-1})$, the multivariate Gaussian distribution
on $\mathbb{R}^p$ with zero mean and precision matrix $\bfTheta$. For
simplicity we focus on the precision matrix. Note that the inverse of its maximum likelihood estimate is a maximum likelihood estimate of the covariance matrix. In this case we have,
\[
\phi(\bfx,\bfTheta) = \frac{1}{2}\,\bfx^\top \bfTheta \bfx
\quad\mbox{with}\quad 
\grad_{\bfTheta} \phi(\bfx,\bfTheta) = \frac{1}{2}\,\bfx\bfx^\top.
\]
If $\bfX=[\bfx_1 \mid \cdots \mid \bfx_n]$ and $\widetilde{\bfX}=[\widetilde{\bfx}_1\mid\cdots \mid \widetilde{\bfx}_N]$ 
are, respectively, sample matrices whose columns are i.i.d from
$N_p(\boldsymbol{0},\bfTheta^{-1}_\mathrm{T})$ and $N_p(\boldsymbol{0},\bfTheta^{-1})$, then
\eqref{eq:ex-gradest} is given by
\[
\widehat{\bfg}(\bfTheta) =  \frac{1}{2n}\,\bfX\bfX^\top  - \frac{1}{2N}\,\widetilde{\bfX}\widetilde{\bfX}^\top. 
\]
That is, $\widehat{\bfg}(\bfTheta) =(\widehat{\bfTheta}_\mathrm{T} - \widehat{\bfTheta})/2$, where $\widehat{\bfTheta}_\mathrm{T}$ and
$\widehat{\bfTheta}$ are, respectively, the maximum likelihood estimates of $\bfTheta_\mathrm{T}$ and $\bfTheta$ based on $\bfX$ and
$\widetilde{\bfX}$. But this gradient approximation requires obtaining the samples $\bfX$ and $\widetilde{\bfX}$. If $\bfC_\mathrm{T}$ and
$\bfC$ are square roots of $\bfTheta_T^{-1}$ and $\bfTheta^{-1}$, then we can use $\bfX = \bfC_\mathrm{T}\bfZ$ and 
$\widetilde{\bfX} = \bfC\widetilde{\bfZ}$, where $\bfZ$ and $\widetilde{\bfZ}$ are matrices of i.i.d $N(0,1)$ variables. But obtaining $\bfC$ at every iteration  is computationally expensive. Langevin dynamics could be used instead to generate the sample matrices but convergence can be very slow. To consider a concrete example, we use the mildly ill-conditioned covariance matrix:
 $$ \bfTheta_{\rm T} = \begin{pmatrix} 1000 & -1 \\ -1 & 2 \end{pmatrix}, $$
 and the Langevin-dynamics iteration (see, e.g., \cite{pml2Book}[p.834])
 \[
  \widetilde \bfX_{k+1} = \widetilde \bfX_k - {\frac {\delta^2}{2}} \bfTheta \widetilde \bfX_k + \delta \,\bfepsilon_k, 
 \]
 where $\bfepsilon_k$ are i.i.d $N_{p\times n}(\boldsymbol{0},\bfI)$.
 Figure \ref{fig:langevin} shows samples obtained after different numbers of iterations.
 The figure clearly demonstrates how ill-conditioning can yield very slow convergence even in 2D. If every SGD step requires such a calculation in high dimensions and with non-trivial functions, then computing the maximum likelihood estimator   becomes infeasible \cite{lin2022on}.
  
\begin{figure}
    \centering
    \begin{tabular}{cc}
    \includegraphics[width=5cm]{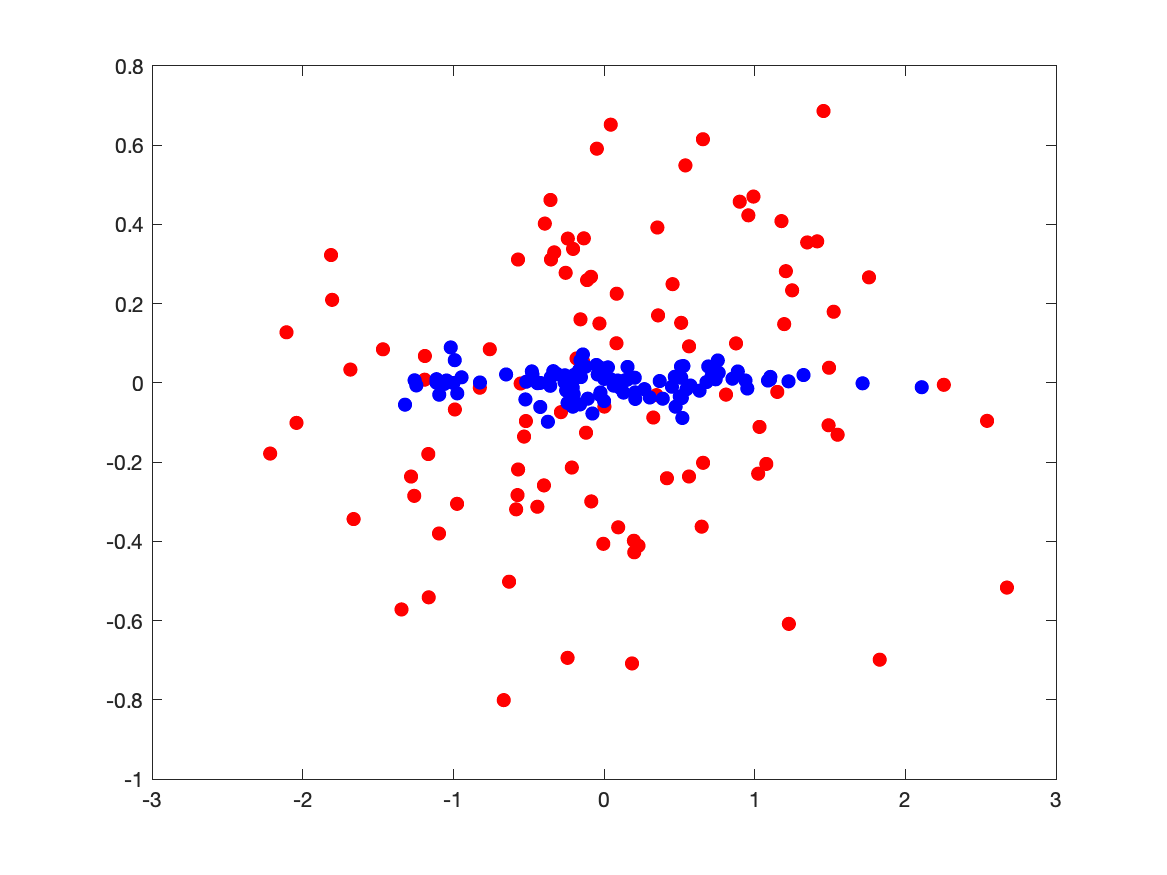} & \includegraphics[width=5cm]{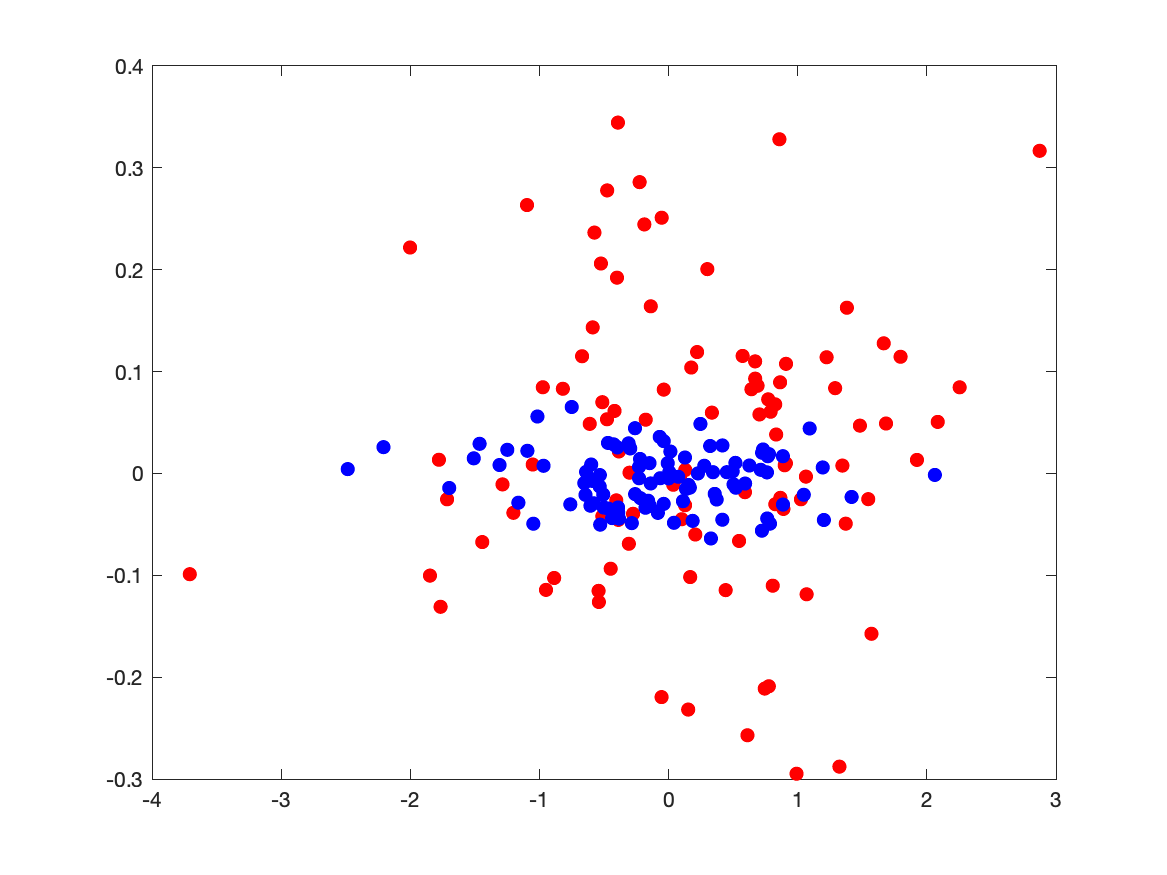} \\
    1000 iterations & 2000 iterations \\
    \includegraphics[width=5cm]{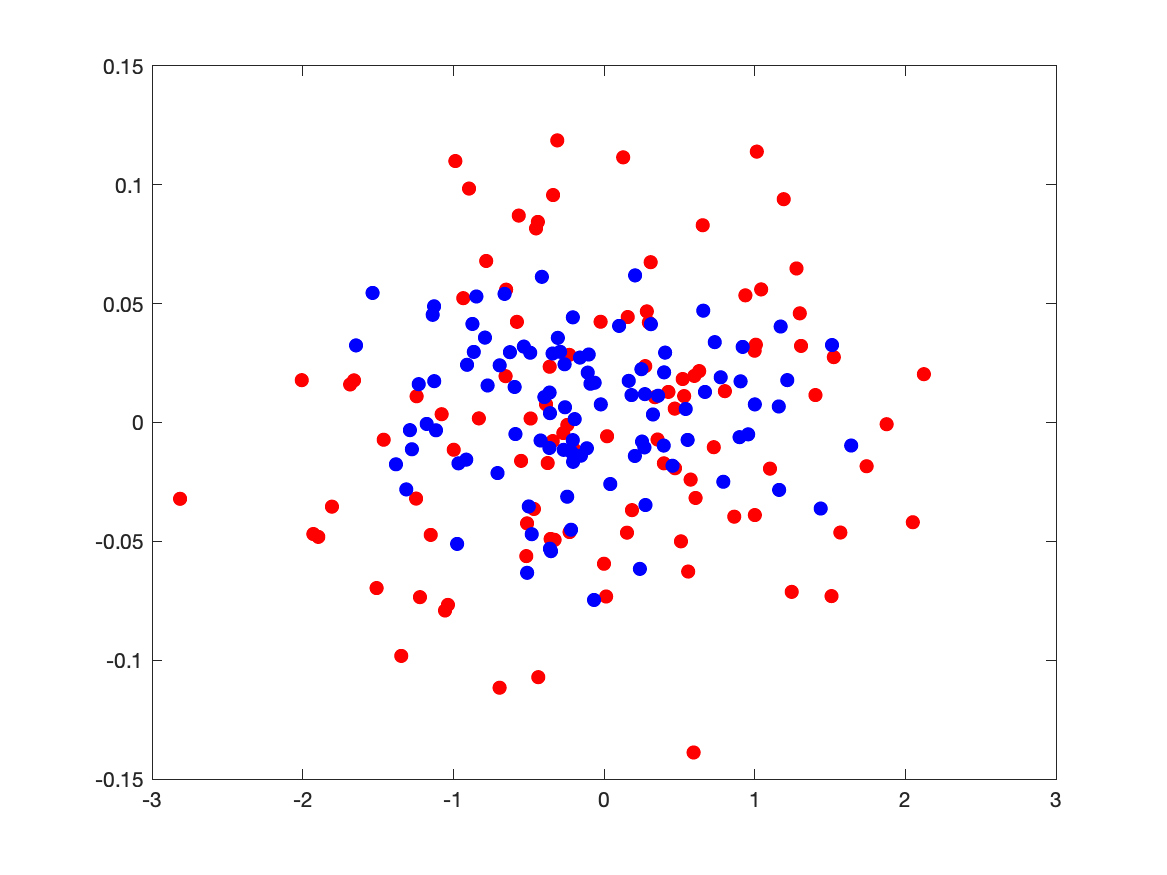} & \includegraphics[width=5cm]{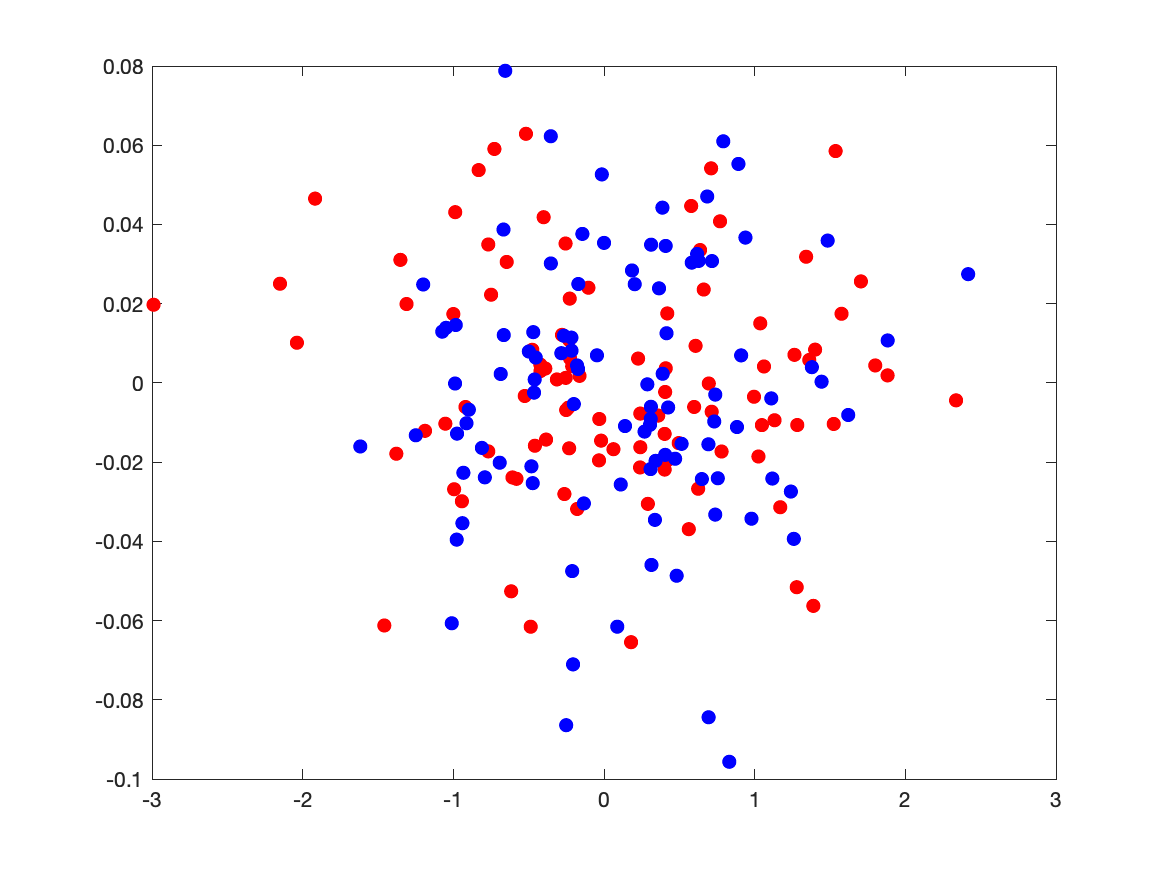} \\
    3000 iterations & 4000 iterations
    \end{tabular}
    \caption{Samples from the true distribution (red) and samples
    obtained with Langevin dynamics (blue) using different numbers of iterations. The figure shows that Langevin dynamics may converge slowly for ill-conditioned problems.}
     
    \label{fig:langevin}
\end{figure}

\end{example}

Example \ref{ex:precision-mat} motivates us to take a different approach to find an estimator of $\bftheta$  that is significantly easier to compute.  
We propose a method that allows the use of standard neural-network techniques without the expensive computation of the partition function or its gradient, and without computationally expensive sampling. To motivate the main idea,
recall the connection between finding a maximum of a posterior distribution (i.e., a MAP) for a Bayesian inverse problem and  finding a solution of a regularized inverse problem (see, e.g., \cite{tenIP}). Choosing a regularization parameter (or a regularization functional) for the latter translates into a tuning of the prior for the former.
The idea of learning the regularization functional is not new, early work can be found in \cite{haten}, where bi-level optimization is used. More recent
work can be found in
\cite{bai2020deep,gonzalez2022solving,meinhardt2017learning, ongie2020deep,ye2018deep}, where proximal mapping methods are used to achieve the same goal in more efficient ways. In particular, the connection between the MAP estimate and regularization has also been exploited in \cite{gonzalez2022solving} to estimate a potential that is used within a proximal iteration.

\medskip

The rest of the paper is organized as follows. In Section \ref{sec2} we define
the Maximum Recovery MAP  estimator $\widehat{\bftheta}$ of the parameters $\bftheta$ of a potential, which is based on the MAP
of the posterior distribution of some latent data given samples from $p_\bftheta$.
This estimator allows us to avoid the partition function and replace sampling from $p_\bftheta$ by solving  an optimization problem.
In Section~\ref{sec3} we consider the more general case where the form of the potential is learned and its parameters are estimated using the method from Section \ref{sec2}. The potential is modeled using a specific form that is analogous to those used in standard variational approaches  in optimal control \cite{BetJ:01}.
This choice  leads to a hyperbolic network \cite{chang2019antisymmetricrnn,chen2019symplectic, 
eliasof2021pde, eliasofmimetic2022,lensink2019fully,ruthotto2020deep} and an optimization
problem whose solution is approximated using a simple neural network with two skip connections.
In Section~\ref{sec4} we summarize the results of numerical experiments that illustrate the performance of our methods. Section \ref{sec5} provides a summary and discussion.
\edits{A proof of strong consistency of the estimator defined in
Section \ref{sec2} is included in Section \ref{sec:consistency}.}

\section{Definition of the  Maximum Recovery MAP Estimator}
\label{sec2}

We start with the case of estimating the parameters of a potential given its known  parametric form. The case when the parametric form is unknown is discussed in Section \ref{sec3}.
 
 Let $\bfX=[\bfx_1 \mid \cdots \mid \bfx_n]$ be the data matrix of i.i.d columns with   density $p_\bftheta$ of the form \eqref{eq:dist}. The goal is to compute an estimate of $\bftheta$. The maximum likelihood estimator is defined
 as a value of $\bftheta$ that makes the data $\bfX$ most likely. But
 we have seen that this is difficult to compute when there is no analytic expression for the partition function, so we ask instead for a value of $\bftheta$ that places $\bfX$ close
 to a mode of the posterior distribution of $\bfX$ given some latent data $\bfd_1,\ldots,\bfd_n$.
 The  latent data are generated as noisy (indirect) observations of the original data $\bfx_i$:
 \begin{eqnarray}
\label{data}
\bfd_i = \bfP_i \bfx_i + \bfepsilon_i\qquad i=1,\ldots,n,
\end{eqnarray}
where $\bfepsilon_i$ are i.i.d  $N_m(\boldsymbol{0},\sigma^2\bfI)$ and
$\bfP_i$ are $m \times p$ ``forward" matrices.  
We use either
$\bfP_i=\bfI$ (noisy observations of $\bfx_i$ as in \cite{gao2021learning}) or choose
$\bfP_i$ to be i.i.d random matrices with 
${\mathbb E}\,\bfP_i^{\top} \bfP_i \propto \bfI$ (we will mostly consider matrices that randomly select a few entries of
$\bfx_i$). Such matrices tend to reduce the information about $\bfx$ in the data and thus increase the relevance of the prior.
Since $(\bfd_i,\bfx_i,\bfP_i)$ are i.i.d, we consider the posterior  
of each $\bfx_i$ separately.
The posterior density of the $\bfx_i$  given $\bfd_i$ and $\bfP_i$ is 
\[
p_\bftheta(\bfx_i\mid\bfd_i,\bfP_i) \propto 
\exp\left(\,-\frac{1}{2\sigma^2}\|\bfP_i\bfx_i - \bfd_i\|^2\, - \phi(\bfx_i,\bftheta)\right).
\]

A maximizer  $\widehat{\bfx}_i(\bftheta,\bfd_i,\bfP_i)$ 
of this posterior distribution (a MAP) is then defined by
\begin{eqnarray}
\label{map} 
\widehat \bfx_i(\bftheta,\bfd_i,\bfP_i) ={\arg\min_{\bfx}}\ {\frac 1{2\sigma^2}} \|\bfP_i \bfx - \bfd_i\|^2 +  \phi(\bfx,\bftheta).
\end{eqnarray}
For simplicity, henceforth we will often write $\widehat{\bfx}_i$ in place
of $\widehat \bfx_i(\bftheta,\bfd_i,\bfP_i)$.
Equation \eqref{map} is equivalent to estimating $\bfx_i$ by minimizing the data misfit $\|\bfP_i\bfx-\bfd_i\|$ with penalty functional $\phi(\bfx,\bftheta)$ and regularization parameter $\bftheta$.
Since we want all the $\bfx_i$ to be close to a mode of their corresponding posteriors,
we define the Maximum Recovery MAP estimator (MR-MAP)  
as the value of $\bftheta$ that minimizes the sum of the mean squared errors
of $\widehat \bfx_i(\bftheta,\bfd_i,\bfP_i)$ as estimators of their corresponding $\bfx_i$ when averaged over $\bfd_i$ and $\bfP_i$ with $\bfx_i$ fixed.
That is, we want to minimize the total frequentist mean squared error (MSE) of the MAP estimates:
\begin{equation}
\label{eq:thestE}
\widehat{\bftheta}_* = \arg\min_{\bftheta}
\medmath{\sum_{i=1}^n} \,\mathbb{E}\left(\,\|\,\widehat \bfx_i-\bfx_i\,\|^2\,\mid\, \bfx_i\,\,\right).
\end{equation}
For the case $\bfP_i\neq \bfI$, we have obtained more stable results when $\bftheta$ is chosen so as to minimize a combination of the MSE and the
predictive MSE:
\begin{equation}
\label{eq:thestPE}
\widehat{\bftheta}_*(\alpha) = \arg\min_{\bftheta}\,\,
\medmath{\sum_{i=1}^n} \,\mathbb{E}\left(\,\,\|\widehat \bfx_i-\bfx_i\,\|^2\,\mid\, \bfx_i\,\,\right) +
\alpha\,\medmath{\sum_{i=1}^n} \,\mathbb{E}\left(\,\,\|\, \bfP_i\widehat{\bfx}_i-\bfP_i\bfx_i\,\|^2\,\mid\, \bfx_i\,\,\right),
\end{equation}
where $\alpha\geq 0$ is a regularization parameter.
Recall that the MSE can be decomposed as a sum of bias and variance
components:
\[
\mathrm{MSE}(\widehat{\bfx}_i)= 
\|\mathbb{E}\,\widehat{\bfx}_i-\bfx_i\|^2
+ \mathrm{trace}(\, \mbox{$\mathbb{V}$ar}(\widehat{\bfx}_i)\,).
\]
Different choices of the matrices $\bfP$, $\sigma$ and $\alpha$ help to balance these components.

 In practice, a sample estimate is used instead of the expectation in \eqref{eq:thestE}. 
 In our implementation we simply minimize the sample averages: 
\begin{eqnarray}
 \widehat{\bftheta} &=& \arg\min_{\bftheta}
\frac{1}{n}\,\medmath{\sum_{i=1}^n}\,\|\, \widehat{\bfx}_i-\bfx_i\,\|^2\label{eq:th-sest}\\
 \widehat{\bftheta}(\alpha) &=& \arg\min_{\bftheta}\,\,
\frac{1}{n}\,\medmath{\sum_{i=1}^n}\,\|\, \widehat{\bfx}_i-\bfx_i\,\|^2
+\frac{\alpha}{n}\,\medmath{\sum_{i=1}^n}\,\|\, \bfP_i\widehat{\bfx}_i-\bfP_i\bfx_i\,\|^2.
\label{eq:th-sPest}
\end{eqnarray}
These two estimates are functions of all the $(\bfx_i,\bfd_i)$ while the estimates \eqref{eq:thestE} and \eqref{eq:thestPE} only depend on the $\bfx_i$. Note that the use of the predictive error $\|\bfP_i\widehat{\bfx}_i-\bfP_i\bfx_i\|^2$
is common in inverse problems. Generalized cross-validation is an estimate
of this error that is often used to choose the regularization parameter in Tikhonov regularization of inverse problem (see, e.g., \cite{tenIP}).

In Bayesian inference the prior is updated using the information provided by the data, and thus the prior should not dominate the likelihood to allow for such update. Our case is different; since the interest is in the prior, it should not be swamped by the likelihood. Our goal is to tune the prior by obtaining good predictions of the prior data $\bfX$ we already have. The likelihood is used to define the way we make such predictions.
 
 Note also that none of the calculations above requires the partition function. We have replaced the evaluation of the partition function with the problem of solving a non-linear optimization problem for $\widehat \bfx(\bftheta,\bfd,\bfP)$ for various forward modeling matrices $\bfP$, latent data $\bfd$ and prior data $\bfx$. We will show that this problem can be solved with an efficient algorithm that allows for the computation of the solution without ever estimating the partition function or its gradients.
While choosing $\bfP = \bfI$ may suffice in some applications, having $\bfP$ chosen as a random projection can significantly improve  the results. To explain why, we note that the estimation of the potential is similar to a different known technique used for generative models. The method can be thought of as an  auto-encoder (AE), where the encoding (latent vector) is achieved by a random matrix $\bfP$ and the decoding is done by our network presented below. The main difference is that while in standard AE both encoders and decoders are optimized, our technique requires the decoder to work well (at least on average) for any random encoder.

\begin{example}\label{ex:var1d}
We consider a one-dimensional version of Example \ref{ex:precision-mat}. In this case, it is simpler to use $\theta>0$ to denote the variance instead of the precision, and $P=1$ is a natural choice. Let $x_1,\ldots,x_n$ be an i.i.d sample from $N(0,\theta)$. The latent data are $d_i=x_i+\epsilon_i$ with
$\epsilon_1,\ldots,\epsilon_n$ i.i.d $N(0,\sigma^2)$. The posterior
distribution of $\bfx=(x_1,\ldots,x_n)$ given 
$\bfd=(d_1,\ldots,d_n)$ is 
$$ p(\bfx|\bfd) \propto \exp \left(-{\frac {\|\bfx-\bfd\|^2}{2\sigma^2}} - {\frac {\|\bfx\|^2}{2\theta}} \right),$$
which defines a 
Gaussian distribution with mean and covariance matrix given by
(general formulas for the Bayesian Gaussian linear model can be found in \cite{tenIP}):
\[
\mathbb{E}\,(\bfx\mid \bfd)= \frac{\theta}{\sigma^2+\theta}\,\bfd,\qquad
\mbox{$\mathbb{V}$ar}\,(\bfx\mid \bfd) = v_\theta\,\bfI,
\qquad v_\theta = \frac{\sigma^2\theta}{\sigma^2+\theta}.
\]
This posterior distribution has a single maximum equal to its posterior
mean, that is
\[
\widehat{\bfx}(\bfd)= \frac{\theta}{\sigma^2+\theta}\,\bfd.
\]
In cases like this, where the posterior distribution is symmetric and the MAP
is the posterior mean, the MAP minimizes the expected MSE (i.e., the Bayes risk for squared-error loss, see, e.g., \cite{schervish}). That is,
for any other estimator $\boldsymbol{\delta}(\bfx)$ with finite variance,  
\[
\mathbb{E}\,\mathrm{MSE}(\,\widehat{\bfx}(\bfd))=\mathbb{E}\,\|\,\widehat{\bfx}(\bfd)-\bfx\,\|^2 \leq 
\mathbb{E}\,\|\,\boldsymbol{\delta}(\bfd)-\bfx\,\|^2=\mathbb{E}\,
\mathrm{MSE}(\,\boldsymbol{\delta}(\bfd)),
\]
\edits{where $\bfd \sim N_n(\bfx,\sigma^2\bfI)$}.
This optimality property follows easily from properties of conditional expectation.
In our case, we do not want to minimize the Bayes risk over all $\theta$ (this just yields $\theta=0$) but the mean squared error of estimating the fixed $x_i$
from a prior with unknown $\theta$. The estimate \eqref{eq:thestE} can be obtained explicitly:
The MSE of $\widehat{\bfx}$ is
\[
\mathrm{MSE}(\,\widehat{\bfx}(\bfd)\,) = \frac{\sigma^4}{(\sigma^2+\theta)^2}\,
\|\bfx\|^2 + \frac{n\,\sigma^2\theta^2}{(\sigma^2 + \theta)^2},
\]
which is minimized with 
\[
\widehat{\theta}_*\, =\, \frac{1}{n}\,\|\bfx\|^2.
\]
This is precisely the method of moments and maximum likelihood estimator
of $\theta$. By the strong law of large numbers,  $\widehat{\theta}_*\,\,\arras\,\,\theta$ as the sample size $n\to\infty$. 
\edits{Under regularity conditions, this strong consistency can be proved when the expected value of the MSE under the prior has a unique minimum at the true value of the parameter
(i.e., assumption (v) in Section \ref{sec:consistency}).}
For the estimate \eqref{eq:th-sest} we assume for simplicity that $n$ is large enough to ignore the constraint $\theta>0$ in the minimization. The solution is then
\[
\widehat{\theta} \,=\, \arg\min_\theta \frac{1}{n}\,
\left\|\,\frac{\theta}{\sigma^2+\theta}\,\bfd - 
\bfx\,\right\|^2\,=\,\frac{\sigma^2\,\bfd^\top\bfx}
{\|\bfd\|^2 -\bfd^\top\bfx},
\]
and since $\|\bfd\|^2/n\,\,\arras\,\,\sigma^2 + \theta$ as $n\to \infty$, we obtain again
\[
\widehat{\theta}\,\,\,\arras\,\,\,\frac{\sigma^2\,\theta}
{\sigma^2 + \theta-\theta}= \theta.
\]

We now compare our method to the one proposed in
\cite{gao2021learning}. In this work $\theta$ is estimated by maximizing a conditional likelihood defined by the posterior distribution of $\bfx$ given $\bfd$, which, as we saw above, is given by
\[
p_\theta(\bfx\mid\bfd)\, \propto\, \frac{1}{v_\theta^{n/2}}\,
\exp\left(\,-\frac{\|\bfx-\widehat{\bfx}(\bfd)\|^2}{2v_\theta}
\,\right).
\]
This is equivalent to solving the minimization
\[
\widetilde{\theta}\, =\, \arg\min_\theta\,\,
 \frac{1}{2\,v_\theta}\,\|\bfx-\widehat{\bfx}(\bfd)\|^2
 + \frac{n}{2}\,\log v_\theta.
\]
Thus, the objective function for $\widetilde{\theta}$ 
is the objective function for $\widehat{\theta}$ scaled  and shifted by functions of the posterior variance $v_\theta$, 
which in turn depends on the partition function.
It easy to show that the minimum is achieved with
\[
\widetilde{\theta}\,=\,
\frac{2\sigma^2\,\|\bfx\|^2 - n\sigma^4 + \sigma^2
\sqrt{n^2\sigma^4 + 4 \|\bfx\|^2 \|\bfd\|^2}}
{2(n\sigma^2 +\|\bfd\|^2 -\|\bfx\|^2)}.
\]
Again, using the strong long of large numbers we find
\[
\widetilde{\theta}\,\,\,\arras\,\,\,
\frac{2\sigma^2\theta -\sigma^4 + \sigma^2
\sqrt{\sigma^4 + 4\theta^2 + 4\sigma^2\theta}}
{4\sigma^2} \,\,=\,\,\theta
\]
as $n\to\infty$. We see that both, $\widehat{\theta}$ and
$\widetilde{\theta}$, are strongly consistent estimates of $\theta$. Also both depend on the choice of $\sigma$ and
the sample size $n$. One obvious advantage of our estimate $\widehat{\theta}$ is that its computation does not require the partition function. 

This example can of course  
be extended to the multivariate problem of estimating
the covariance matrix of a multivariate Gaussian
distribution.
Let $\bfX=[\bfx_1\mid \cdots\,\mid\bfx_n]$, where
the columns are i.i.d $N_p(\boldsymbol{0},\bfTheta)$.
Again, the maximum likelihood estimate of $\bfTheta$ is
$\bfX\bfX^\top/n$.
To determine an estimate using \eqref{eq:thestE}, we generate $\bfd_i$ as in \eqref{data}. The conditional distribution of $\bfd_i$ given
$\bfx_i$ and $\bfP_i$ is then $N_m(\bfP_i\bfx_i,\sigma^2\,\bfI)$. As in the one-dimensional case, the posterior of $\bfx_i$ given $\bfd_i$ and $\bfP_i$ is Gaussian with posterior mean equal to its MAP:
\[
\widehat{\bfx}_i =
(\,\bfP_i^\top\bfP_i +\sigma^2\,\bfTheta^{-1}\,)^{-1}\bfP_i^\top\bfd_i.
\]
For simplicity, set 
$\bfH_i(\bfTheta) = (\,\bfP_i^\top\bfP_i +\sigma^2\,\bfTheta^{-1}\,)^{-1}$.
Since the bias and covariance matrix are 
\begin{eqnarray*}
\mathrm{Bias}(\,\widehat{\bfx}_i\mid \bfx_i,\bfP_i\,) &=& -\sigma^2\,\bfH_i(\bfTheta)\,\bfTheta^{-1}\bfx_i\\
\mathbf{V}\mbox{ar}(\,\widehat{\bfx}_i\mid\bfd_i,\bfP_i) &=& \sigma^2\,\bfH_i(\bfTheta)\bfP_i^\top \bfP_i
\bfH_i(\bfTheta),
\end{eqnarray*}
we see that \eqref{eq:thestE} reduces to
\[
\widehat{\bfTheta}_* = \arg\min_{\bfTheta}\,\,
\sigma^4\,\medmath{\sum_i }\,\mathbb{E}(\,\|\bfH_i(\bfTheta)\bfTheta^{-1}\bfx_i\|^2\,\mid\,\bfx_i\,)
+\sigma^2\,\medmath{\sum_i}\,\mathbb{E}\,\mathrm{trace}( \bfH_i(\bfTheta)^2\bfP_i^\top\bfP_i\,).
\]
In the case $\bfP_i = \bfI$, it is straight forward to show that we again recover the maximum likelihood estimate, $\widehat{\bfTheta}_* = \bfX \bfX^{\top}/n$, \edits{which in this case is also strongly consistent.}
\end{example}

 \medskip
 \edits{Showing consistency of the estimates in Example \ref{ex:var1d} was straightforward but consistency is not guaranteed in general, it depends on, among other factors, the likelihood used with the given prior. In Example \ref{ex:var1d} the Gaussian likelihood was appropriate for the prior but it 
 cannot be appropriate for every prior.
 For example, a Laplace prior with $\phi(x,\theta)=-\theta\, |x|$, $\theta>0$, coupled with an analogous Gaussian
 likelihood yields an inconsistent estimator (in this case
 assumption (v) in Section \ref{sec:consistency} is not satisfied).
Sufficient conditions for consistency are presented in Section \ref{sec:consistency}.  In our discussion we have used Gaussian likelihoods because of the model for
$\phi$ used in Sections \ref{sec3} and \ref{sec4}.
 }
 
\section{Numerical Solution and Optimization}
\label{sec3}

In Section \ref{sec2} we described a way to determine the parameters of a given potential $\phi$. We now consider the more general case when the form of the potential is unknown and is modeled
using a least-action potential that leads to a neural network.
 The weights of the network are then the parameters of the potential that are estimated as explained in Section \ref{sec2}.

 \subsection{Motivation}
\edits{One possible approach to build the potential $\phi$ is to find a representation using a rich class of functions; for example, we could use a deep neural network.
 However, this approach typically leads to highly non-linear and non-convex problems. As a result, even if we are given an optimal set of parameters for the network, it is difficult to sample from a distribution with such a potential. Furthermore, since training the network requires solving the optimization problem \eqref{map},  estimating this potential can be difficult.  One way to obtain a convex yet very expressive potential is to increase the dimensionality of the problem and to consider each layer of the network as a part of the potential. A recent  approach to neural networks discussed in \cite{HaberRuthotto2017} is to view them as Ordinary Differential Equations (ODE) or as dynamical systems that take an original set of points (the data) and transform them into a different configuration to fulfill a particular task. In this spirit, a potential can be thought of as the generator of this ODE, that is, the dynamics. This idea is common in mathematical physics where the Euler-Lagrange equations minimize some energy that leads to a particular dynamics. This approach has also been  used to design neural networks with desired properties such as feature reversibility \cite{eliasofmimetic2022}, and energy preservation \cite{eliasof2021pde}. The energy is typically defined as the sum of potential and kinetic energies. As we show next, we build our network from similar components, mimicking the behavior of natural potentials using neural networks, which results in a convex and expressive potential in high dimensions.}
 
 To be specific, recall that two key ingredients of neural networks are: (i) the high-dimensional representation $\bfu_0 \in \mathcal{U}$ 
 (often called an `embedding') 
 of the data $\bfx \in \mathcal{X}$, and (ii) a sequence
of outputs $[\bfu_1, \ldots , \bfu_{\ell}]$ of an $\ell$-layer neural network that operates in $\mathcal{U}$.
For simplicity we assume $\mathcal{X}=\mathbb{R}^p$ and $\mathcal{U}=\mathbb{R}^q$
with $q>p$, such that the higher-dimensional representation of the data, $\bfu_0$, is then obtained by multiplication with a $p\times q$ learnable matrix $\bfK$. 
We regard to the sequence of outputs $[\bfu_0, \ldots , \bfu_{\ell}]$ as a \emph{flow} that defines
a vector with $q \times \ell$ entries corresponding to the given data point $\bfx$: $\bfu = [\bfu_0^{\top}, \bfu_1^{\top}, \ldots, \bfu_{\ell}^{\top}]^{\top}$.
The last vector $\bfu_\ell$ plays a particular role as it is the last outcome of the flow, and it can be obtained
from $\bfu$ by multiplication with the matrix
 $\bfQ_\ell=[\,\boldsymbol{0},\ldots,\boldsymbol{0}, \bfI\,]$.

It what follows, we will find a density for $\bfu$ of the form
$p_\bftheta(\bfu) \propto \exp(-\phi(\bfu,\bftheta))$ for
some non-negative potential  $\phi$ and parameters $\bftheta$,
which in turn provides a distribution on $\mathbb{R}^q$ defined
by the distribution of $\bfu_\ell = \bfQ_\ell\, \bfu$.
Thus, we determine a density for $\bfu$ instead of $\bfx$.

\subsection{Network Architecture}
 \label{sec:net-arch}

We start by describing the network architecture used to define a potential $\phi(\bfu, \bftheta)$, based on an initial data point $\bfx$.

First, note that a MAP defined by \eqref{map} for given number of layers 
$\ell$, observed data $\bfd=\bfP\bfx + \bfepsilon$, projection operator $\bfP$, and learnable weights $\bftheta$ is given by
\begin{eqnarray}
\label{mapu} 
\widehat{\bfu} =\arg\min_{\bfu}\ {\frac 1{2\sigma^2}} \|\bfP \bfK \bfQ_\ell\,\bfu - \bfd\|^2 +  \phi(\bfu,\bftheta).
\end{eqnarray}
Although $\widehat{\bfu}(\bfx)$ is a function of the input
$\bfx$, for brevity, we will use the notation 
$\widehat{\bfu}$. 
 Our high-dimensional reformulation of the problem
 is analogous to the search of
 sparse solutions of inverse problems, where  the regularization function is the $\ell_1$ norm, $\phi(\bfu) = \|\bfu\|_1$, and the matrix $\bfK$ is a dictionary.
As we see next, in our case, we define a $\phi$ whose derivative is a $\ell$-layer neural network.

While it is a common practice to use a neural network with $\ell$ layers to model a potential \cite{bai2020deep,li2016learning,proxDnonoho2018}, such an approach has a few major disadvantages.
First, if $\phi$ is a neural network, then \edits{to solve problem \eqref{mapu}, we need to compute its derivative, which requires an additional} back-propagation and thus two passes through the network are needed to compute
$\grad_{\bfu} \phi(\bfu, \bftheta)$. Moreover,  \eqref{mapu} requires the solution of a non-linear nonconvex optimization problem that is often obtained using proximal iteration \cite{Boyd,li2016learning,proxDnonoho2018, e2e2021}, 
a method similar to those used in ``learning to optimize" \cite{heaton2022explainable,li2016learning,mckenzie2023faster}, that are a special form of recurrent neural networks (RNN) \cite{MartensSutskever2012}. These methods use a forward pass through a network at each iteration  to approximately solve the optimization problem. If a small number of RNN steps are taken, one does not solve the optimization
problem in \eqref{map}, even approximately. Thus, it leads to a degradation of the method and its performance (see \cite{l2o2021} for further discussion). If on the other hand, one uses many steps
of the proximal method, then the computational cost of the problem can be overbearing.

To avoid these difficulties, we model $\phi$ as follows: given a sequence $\bfu_0, \ldots, \bfu_{\ell}$ where $\bfu_i \in {\mathbb{R}}^q$ corresponding to $\bfx\in {\mathbb{R}}^p$, with $p \le q$, we define 
\begin{eqnarray}
\label{potential}
\phi(\bfu, \bftheta) = \phi(\bfu_0, \ldots, \bfu_{\ell}, \bftheta) = {\frac 12} \,\medmath{\sum_{j=1}^\ell}\,\|\bfu_{j} - \bfu_{j-1} \|^2 + 
{h^2}\,\medmath{\sum_{j=0}^{\ell-1}}\, \bfw_j^{\top} f(\bfu_j, \bftheta_j) + \bfr^{\top} \bfu_{\ell},
\end{eqnarray}
where $f$ is a learnable pointwise non-linear function parameterized by $\bftheta$ (to be discussed next), 
$h$ is a hyperparameter to be chosen, and $\bfr \in {\mathbb{R}}^q$  and
$\bfw_j\geq \boldsymbol{0}, \bfw_j \in {\mathbb{R}}^q$ are trainable weights.
Note that this potential is analogous to a discretization of the energy-type integral
$$ 
\Phi(\bfu, \bftheta,T) = \hf \,\medmath{\int_0^T}\,  \|\dot \bfu(t)\|^2\,dt + \medmath{\int_0^T}\,\bfw(t) f(\bfu(t), \bftheta(t))\, dt + \bfr^{\top} \bfu(T), 
$$
where the first term is discretized using the midpoint rule and the second one is discretized using the (left) rectangular rule. Such integrals and energies are standard in optimal control and Lagrangian dynamics.
The first term can be thought of as kinetic energy and the second term as potential energy. By choosing $f(\cdot, \cdot)$ to be a convex function, the obtained potential is also convex. Thus, the optimization problem \eqref{map} can be solved in a straightforward way, yielding a unique solution.

With definition \eqref{potential} of the potential, the optimization problem \eqref{mapu} can be explicitly written as 
\begin{eqnarray}
\label{mapuexp} 
\min_{\bfu}\  \hf\|\bfP \bfK\bfQ_\ell \,\bfu - \bfd\|^2 +  \beta \,\medmath{\sum_{j=1}^{\ell}}\, \hf\|\bfu_{j} - \bfu_{j-1} \|^2 + \beta h^2\,\medmath{\sum_{j=0}^{\ell-1}}\,\bfw_j^{\top} f(\bfu_j, \bftheta_j) + {\beta}\bfr^{\top} \bfu_{\ell},
\end{eqnarray}
where the hyperparameter $\beta$ is a regularization parameter used to balance the data fitting term and the energy.
An important property of this energy-type definition of $\phi$ is that \eqref{mapu} can be solved explicitly once $\bfu_1$ and $\bfu_{0}$ are
initialized.
This is in sharp contrast to the methods presented in \cite{proxDnonoho2018,e2e2021} that require solving a non-linear optimization problem at each step.

Our neural network is then defined by  the derivative of the potential:
differentiating \eqref{mapuexp} with respect to each $\bfu_j$ leads to
\begin{subequations}
\label{dmapu}
\begin{eqnarray}
\label{dmapu1} 
&& \bfu_{j+1} =  2\bfu_j - \bfu_{j-1} +  \grad_{\bfu} f(\bfu_j, \bftheta_j)^{\top} \bfw_j \quad \qquad j = 1, \ldots, \ell-1 \\
\label{dmapu2} 
&& (\bfK^{\top} \bfP^{\top}\bfP\bfK  + \beta \bfI)\, \bfu_{\ell} =  \bfK^{\top} \bfP^{\top} \bfd  +  \beta\, \bfu_{\ell-1} + {\beta}\bfr. 
\end{eqnarray}
\end{subequations}
The system \eqref{dmapu} is a discrete {\em non-linear boundary value problem} where given $\bfu_0$ and condition \eqref{dmapu2}, one can solve for $\bfu_1,\ldots,\bfu_{\ell}$.
Since the problem is a boundary value problem, it has no closed form solution without imposing further assumptions.
One classical method to solve such a system uses shooting \cite{ascher1981collocation, bock1983recent}. In this approach, one chooses $\bfu_0$, takes an initial guess of $\bfu_1$, 
and then propagates the network \eqref{dmapu2} forward to compute the trajectory and obtain some $\widetilde \bfu_{\ell}$ and $\widetilde \bfu_{\ell-1}$. One can then check if 
\eqref{dmapu2} is satisfied with these 
$\widetilde \bfu_{\ell}$ and $\widetilde \bfu_{\ell-1}$.
If it is, then a solution, $\bfu$ (the flow) is found. Otherwise, one needs 
to change $\bfu_1$ so that equation \eqref{dmapu2} approximately holds. Typical ways to change $\bfu_1$ involve either a fixed point  or Newton's method (see \cite{bock1983recent} for a complete discussion).
While we could employ such an approach, it adds further complexity and computational costs that we prefer to avoid. To this end, we assume that the mapping from the final condition to the initial one 
and back is smooth enough to approximate it with a single layer neural network. We then find an efficient way to  solve for $\bfu_1$ using the data at hand.

We therefore first choose $\bfu_0$. In principle, we are free to choose $\bfu_0$ as we please as long as $\bfu_0$ is different (with high probability) for different inputs $\bfx's$. This allows for a unique trajectory (flow) from initial to final condition.
Since we would like to have a short trajectory, we choose  $\bfu_0$ so that it approximately fits the data, that is, 
\begin{eqnarray}
\label{u0}
(\bfK^{\top} \bfP^{\top}\bfP\bfK  + \beta \bfI) \bfu_{0} =  \bfK^{\top} \bfP^{\top} \bfP \bfx,
\end{eqnarray}
where $\beta$ is the regularization parameter in \eqref{mapuexp}. Since the sampling matrix $\bfP$ is independent of the inputs $\bfx$'s
and the latter have a density with respect to the Lebesgue measure, it follows that
$\bfu_0(\bfx) \neq \bfu_0(\bfx')$ with probability
one when $\bfx$ and $\bfx'$ are independent.

However, as explained above, choosing $\bfu_0$ without an appropriate $\bfu_1$ leads to a non-linear boundary value problem that we would like to avoid. To approximate the boundary value problem with an initial value problem we proceed as follows: 
for a  
given $\bfu_0$, we set $\bfu_1$ as a single layer neural network of $\bfu_0$
\begin{eqnarray}
\label{initialization1}
\bfu_{1} = \bfg(\bfu_0, \bfomega),
\end{eqnarray}
where $\bfomega$ are trainable parameters. As we show in the next subsection, the parameters $\bfomega$ are trained using the data so that \eqref{dmapu2} approximately holds. This ensures that no non-linear boundary value problem needs to be solved and that we can propagate {\em only} forward in order to solve \eqref{dmapu}. The learned mapping $\bfg$ yields a $\bfu_1$ that propagates forward to solutions $\bfu_{\ell}$ and $\bfu_{\ell-1}$ that approximately satisfy \eqref{dmapu2}.

A few comments are now in order. First, 
note that the forward propagation is a residual network with a so-called double skip connection \cite{chen2019symplectic,ruthotto2020deep}.
Second,
for small-scale problems it is possible to start by solving \eqref{u0} to then solve \eqref{dmapu2} directly. For large-scale problems one can use  Conjugate Gradient Least Squares (CGLS) \cite{hansen} to efficiently solve  \eqref{u0}, or use a few iterations of Basis Pursuit \cite{chen96atomic}. 
We summarize the network used to solve  \eqref{dmapu1} and \eqref{dmapu2} in Algorithm~\ref{algForu}.

Third, it is important to note that the network defined by \eqref{dmapu1} uses a non-linear function of the form 
$\grad_{\bfu} f(\bfu_j, \bftheta_j)^{\top} \bfw_j.$
It is common to use network parameters $\bfK_j$
and $\bfb_j$ and set
$$ \bfw_j^{\top} f(\bfu_j, \bfK_j, \bfb_j) = \bfw_j^{\top} f(\bfK_j \bfu_j + \bfb_j), $$
whose derivative is given by
$$ \grad_\bfu f(\bfu_j, \bfK_j, \bfb_j)^{\top} \bfw_j = \bfK_j^{\top} (f'(\bfK_j \bfu_j + \bfb_j) \odot \bfw_j). $$
Here, $f'$ is the derivative of the (pointwise) activation function. Choosing $f$ as the integral of some known activation function leads to a simple ``standard" network in \eqref{dmapu1}.
For example, we choose 
$$ f(t) = \hf {\rm max}(t,0)^2 $$
to obtain the usual ReLU activation function in the network.

Our last comment regards with the choice of energy-type potential and network.
First, note that the energy in Equation \eqref{potential} consists of three terms. The first term can be thought of as kinetic energy. It is minimized when the
solution does not change from its initial state. The second term can be seen as a per-step cost or
potential energy, and the third is a linear term that represents a bias. 
For the energy to be small, the network needs to choose an appropriate initial representation $\bfu_0$ of $\bfx$,
so that $\bfu_0$ not only fits the data $\bfd$, but also gives $\bfK\bfu_0 \approx \bfx$. The network  is then optimized to learn weights to correct this initialization.
The energy is large if a significant correction  is needed and small if only a minimal change is required. 

\begin{algorithm} 
\begin{algorithmic}
\caption{Network for the solution of \eqref{dmapu1} and \eqref{dmapu2}.}
\label{algForu}
\STATE $\bfd = \bfP \bfx + \bfepsilon$
\STATE $ \bfu_{0} =  (\bfK^{\top} \bfP^{\top}\bfP\bfK  + \beta \bfI)^{-1}\bfK^{\top} \bfP^{\top} \bfd$
\STATE $\bfu_1 = \bfg(\bfu_0, \bfomega)$
\FOR{$j=1,2,..., \ell-1$}
\STATE $\bfu_{j+1} =  2\bfu_j - \bfu_{j-1} +  \grad_{\bfu} f(\bfu_j, \bftheta_j)^{\top} \bfw_j$
\ENDFOR
\STATE Set $ \bfq = (\bfK^{\top} \bfP^{\top}\bfP\bfK  + \textcolor{blue}{\beta} \bfI)^{-1} (\bfK^{\top} \bfP^{\top} \bfd  +  \textcolor{blue}{\beta} \bfu_{\ell-1} + \textcolor{blue}{\beta}\bfr) $
\STATE return $\bfu_{\ell}, \bfq$
\end{algorithmic}
\end{algorithm}

\subsection{Training the network}
\label{sec:training}
\begin{figure}
    \centering
    \includegraphics[width=1\linewidth]{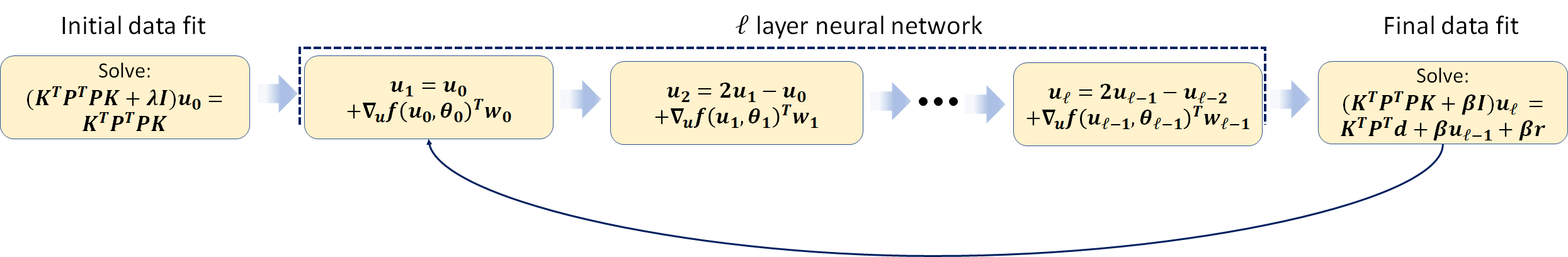}
    \caption{\edits{The overall flow of our approach. We start by solving for an initial solution $\bfu_0$, followed by an iterative application of an $\ell$-neural network and data fitting steps, as described in Equation \eqref{dmapu}.}}
    \label{fig:architecture}
\end{figure}

Training the network requires considering the optimization problems  \eqref{mapuexp} by solving \eqref{dmapu}.
 To solve \eqref{dmapu} given a data point $\bfx_i$, \edits{at every batch iteration,} we choose a random matrix $\bfP_i$ and random noise $\bfepsilon_i$ to get the latent data $\bfd_i =\bfP_i\bfx_i + \bfepsilon_i$.
We recover a corresponding $\widehat{\bfu}_i=\widehat{\bfu}(\bfx_i)$ (note that $\widehat{\bfu}_i$ does not correspond to the i-th layer but rather to the i-th datum) and the corresponding vector $\bfq_i$   defined in Algorithm~\ref{algForu}, and
then compute the loss defined by adding three error terms:
\begin{itemize}
    \item[] Recovery error:
    $R_e = {\frac 1n}\|\bfK\bfQ_\ell\,\widehat \bfu_i - \bfx_i\|^2$
    \item[] Predictive error: $R_p = {\frac 1n}\|\bfP_i
    \bfK\bfQ_\ell\,\widehat \bfu_i- \bfP_i\bfx_i\|^2$
    \item[] Consistency error $R_c = {\frac 1n}\|\bfq_i - \bfQ_\ell \,\widehat \bfu_i\|^2$
\end{itemize} 
\smallskip
Leading to a total loss function of:
\begin{equation}
    \label{eq:total_loss}
    \mathcal{L} = R_e + \alpha R_p + \gamma R_c.
\end{equation}
\edits{While the different objective terms could be scaled, we found that simple addition  of the terms (i.e., $\alpha=\gamma=1$) yields acceptable results.} Note that the recovery error measures how well we recover the data $\bfx_i$ with the MAP based on its noisy sample $\bfd_i$.
The predictive error term measures how well the projection 
$\bfP_i\bfK\bfQ_\ell\widehat{\bfu}_i$ of $\bfK\bfQ_\ell\widehat{\bfu}_i$ approximates the projection $\bfP_i\bfx_i$ of $\bfx_i$.
 Finally,  the third term, the consistency error, learns the initialization of the network in a way that generates outputs consistent with Equation~\eqref{dmapu2}; that is, a $\bfu_{\ell}$ that results from the network \eqref{dmapu1} also approximately  solves \eqref{dmapu2}.
Clearly, if $R_e$ is small then so is $R_p$ but controlling 
$R_p$ ensures that the difference between 
$\bfK\bfQ_\ell\,\widehat \bfu_i$ and $\bfx_i$ is small
when projected with $\bfP_i$. We have found empirically that: (i) optimizing
both $R_e$ and $R_p$ helps convergence; (ii) the consistency term $R_c$ tends to be very small, and (iii) $R_c$ converges quickly, after several training epochs, as depicted from 
\cref{fig:cifar10_consistecyloss}.

Given the loss function, the next step is to use it to update $\bftheta$ and $\bfw$ via backpropagation. \edits{Specifically, we use the Adam  optimizer \cite{adamOptimizer} with a learning rate of $10^{-3}$ and reduce the learning rate by a factor of 0.5 after every 20 epochs for a total of 120 epochs. We also use weight decay regularization with a coefficient of $10^{-5}$.} The overall training details
are summarized in Algorithm \ref{algo:alg2}.
There are a number of free parameters one needs to choose when training this algorithm, such as the network depth (5 in our experiments), the embedding dimension (128 in our experiments) and the regularization parameters. These hyperparameters are typically  tuned in a cross-validation process.

We also note that the network architecture contains two main blocks, as seen in \cref{fig:architecture}. The ``solver" block \eqref{dmapu2}
of fitting the data and the network block \eqref{dmapu1} that ``corrects" the solution so that its corresponding value of the potential
$\phi$ is small.
The data-fitting block is approximated using conjugate gradient \cite{hansen}. \edits{It takes many iterations for $\bfu$ to be approximately a stationary point but in practice we found that using a small number (about five to eight conjugate gradient iterations) was sufficient for training the network.}
Finally, we have found experimentally that training the network for different random $\bfP_i$ at each iteration yields better results, but of course the training takes more time than when using  $\bfP_i=\bfI$. As it has been observed in text processing \cite{baevski2022data2vec}, recovering an image after randomly removing some of its parts seems to yield potentials with better generalization properties.

\begin{algorithm}
\begin{algorithmic}
\caption{Training potentials.}
\label{algo:alg2}
\FOR{$i=1,2,...$}
\STATE Generate random matrix $\bfP_is$ and noise $\bfepsilon_i$
\STATE Compute data $\bfd_i = \bfP_i \bfx_i + \bfepsilon_i$
\STATE Use Algorithm~\ref{algForu} to compute $\bfu_i$ and $\bfq_i$.
\STATE Compute the recovery loss $R_e$, the predictive loss $R_p$ and the consistency loss $R_c$ 
\STATE Set $\mathcal{L} = R_e + \alpha R_p + \gamma R_c$
\STATE Compute the gradient of the loss $\mathcal{L}$ and update $\bftheta$ and $\bfomega$.
\ENDFOR
\end{algorithmic}
\end{algorithm}

\section{Numerical Experiments}\label{sec4}
The algorithms and techniques presented above are now evaluated using a number of examples. We start with an elementary but illustrative experiment where our approach is used to learn a true unknown potential. In addition, we include image recovery examples that are relevant in energy and score-based methods \cite{gao2021learning,song2021scorebased,zhao2016energy}.

\subsection{Datasets and settings}
In addition to the Gaussian mixture potential used in Section \ref{sec:potential_recovery_experiment}, we employ two popular datasets. Namely, CIFAR-10 \cite{krizhevsky2009learning} which consists of a total of 60,000 images from 10 classes, where 50,000 are used for training and 10,000 for testing, and Celeb-A \cite{liu2015faceattributes} which contains 202,599 celebrity face images. We follow the standard training/validation/testing split from \cite{liu2015faceattributes}.
Our method is implemented in PyTorch \cite{paszke2019pytorch}, trained and evaluated using an Nvidia RTX-3090 with a total of 24GB of memory.

\subsection{Potential Recovery}
\label{sec:potential_recovery_experiment}
\begin{figure}
\begin{tabular}{ccc}
\includegraphics[width=4.75cm]{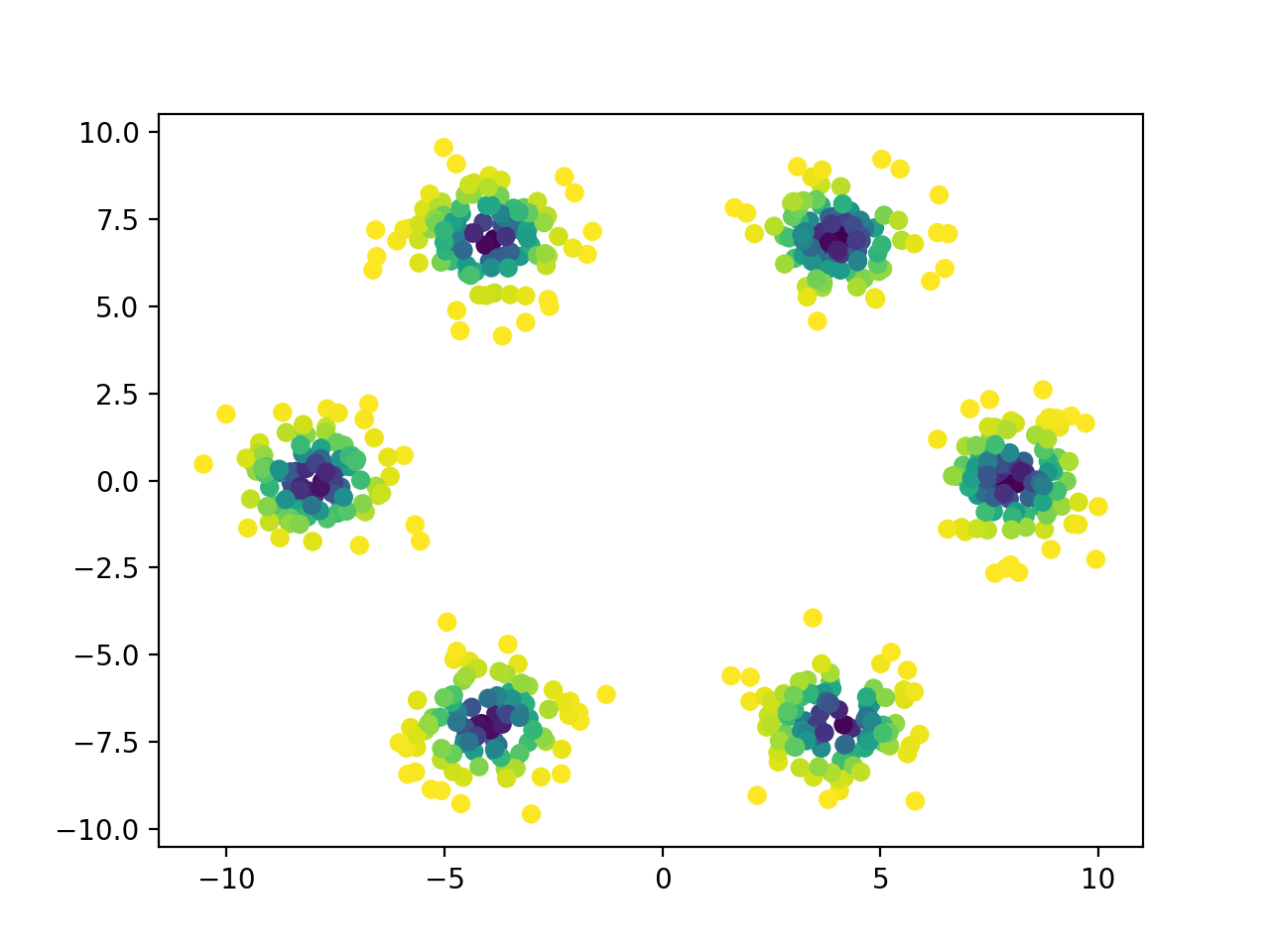} &
\includegraphics[width=4.75cm]{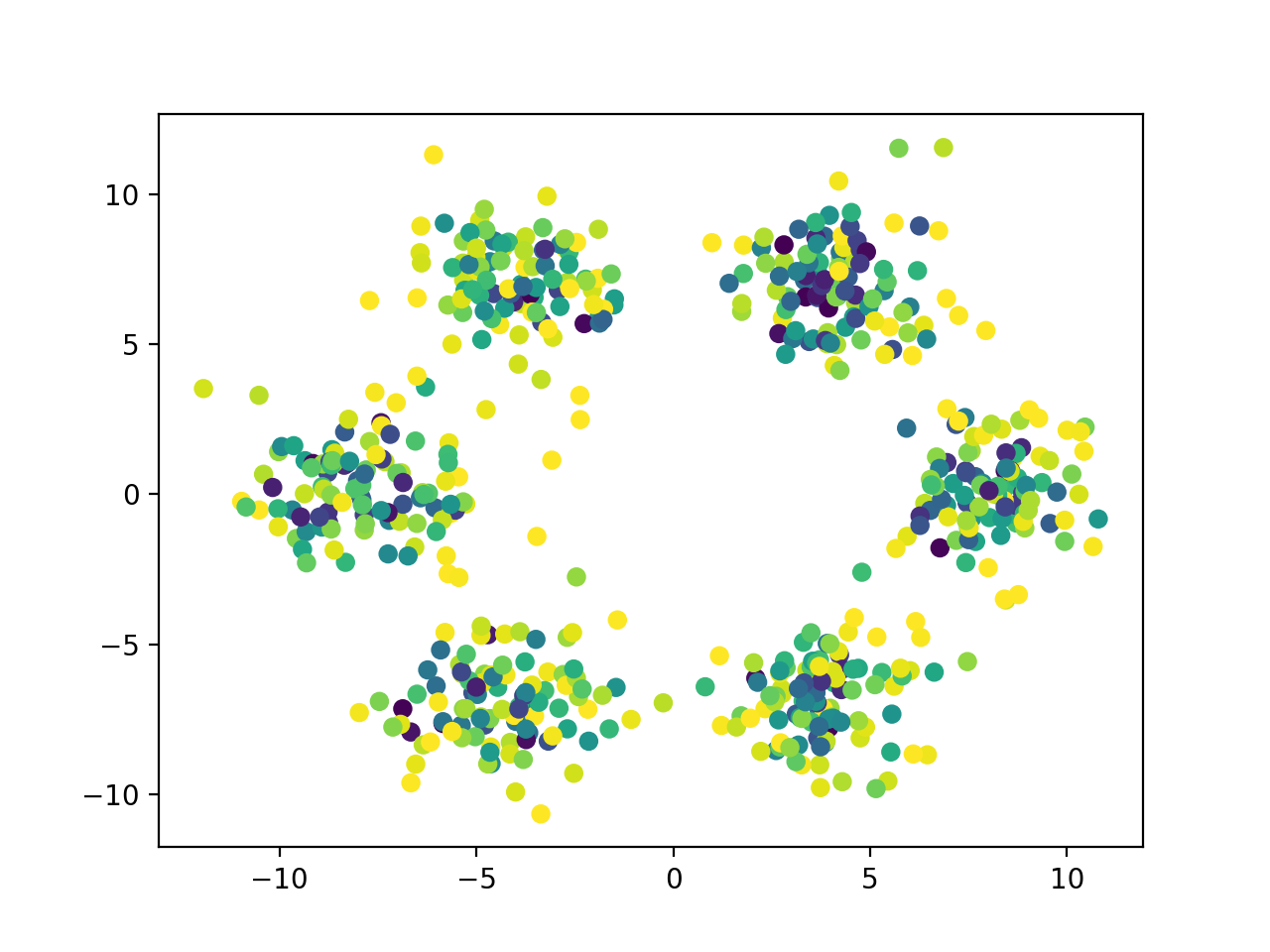} &
\includegraphics[width=4.75cm]{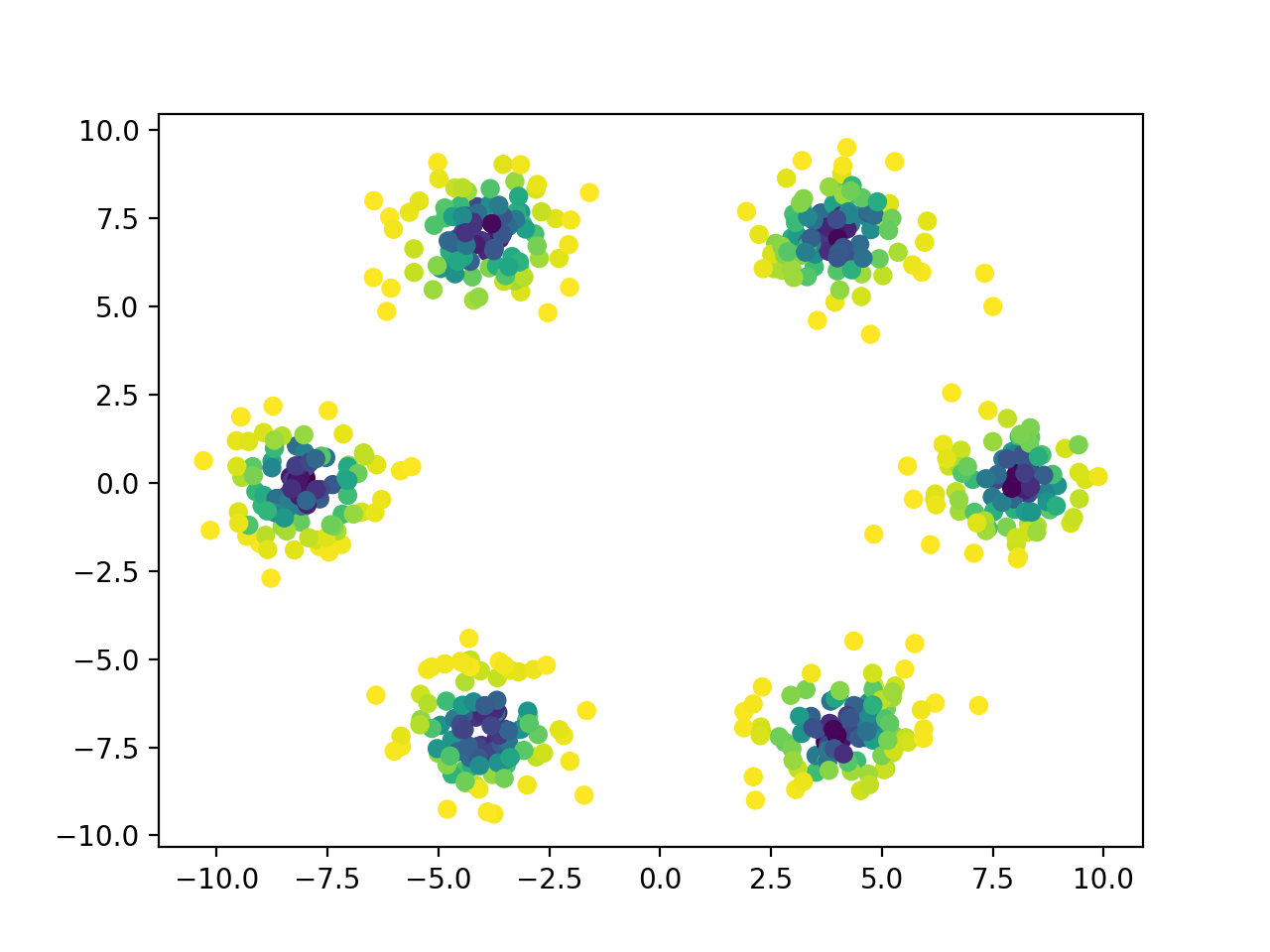} \\
(a) Initial $\bfx_i$   & (b) Latent data $\bfd_i$  & (c) Recovered $\widehat{\bfx}_i$
\end{tabular}
\caption{ (a) Training data $\bfx_i$ used for the Gaussian mixture example. (b) Latent data $\bfd_i=\bfx_i+ \bfepsilon_i$. (c) Recovered
$\bfx_i$ defined as $\widehat{\bfx}_i= \bfK \bfQ_\ell\widehat{\bfu}(\bfx_i)$.
The color of each point represents the value of its potential, 
darker colors correspond to smaller values of the potential.
\label{figdata2d}}
\end{figure}

\begin{figure}
\begin{tabular}{ccc}
\includegraphics[width=4.75cm]{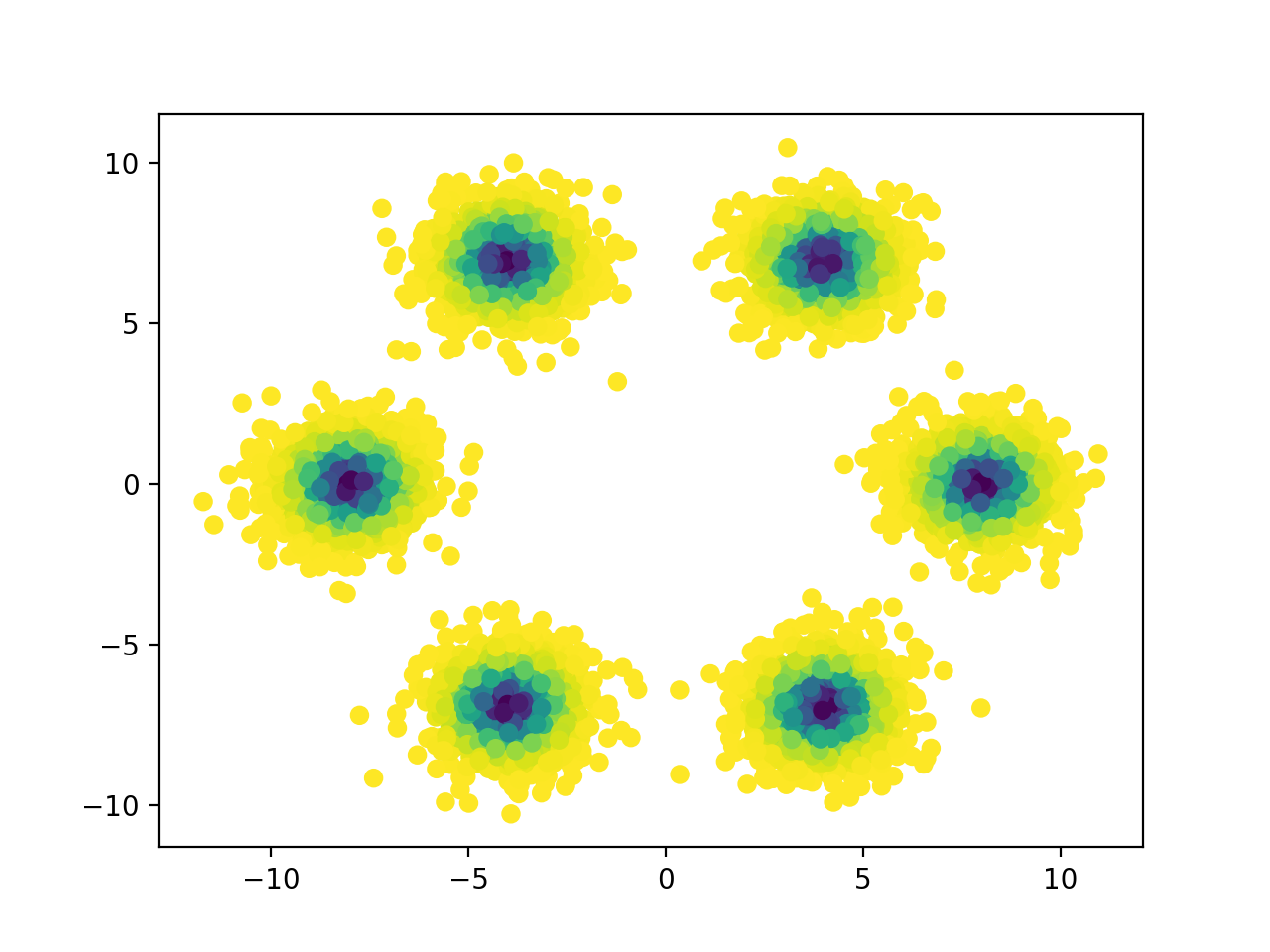} &
\includegraphics[width=4.75cm]{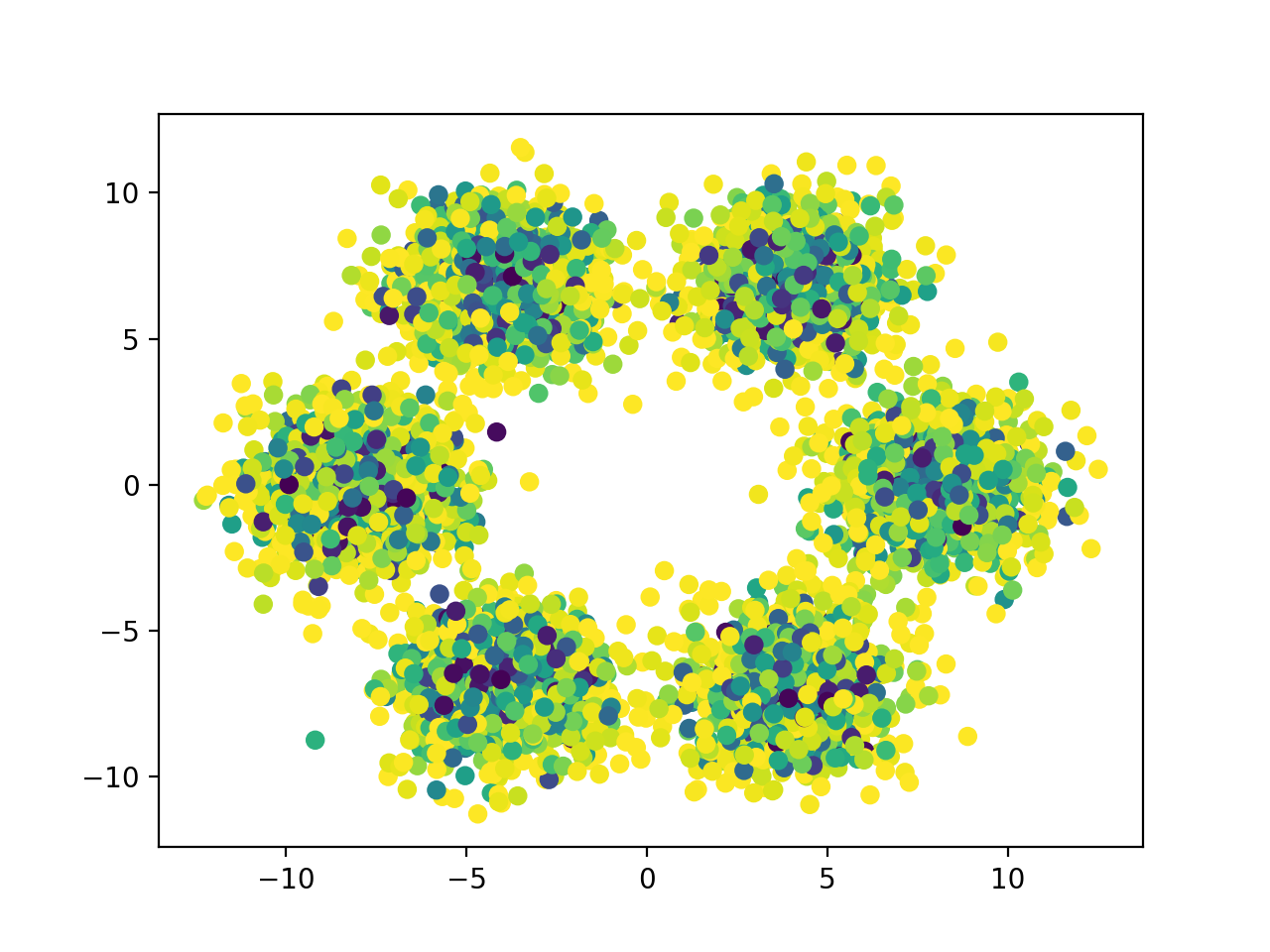} &
\includegraphics[width=4.75cm]{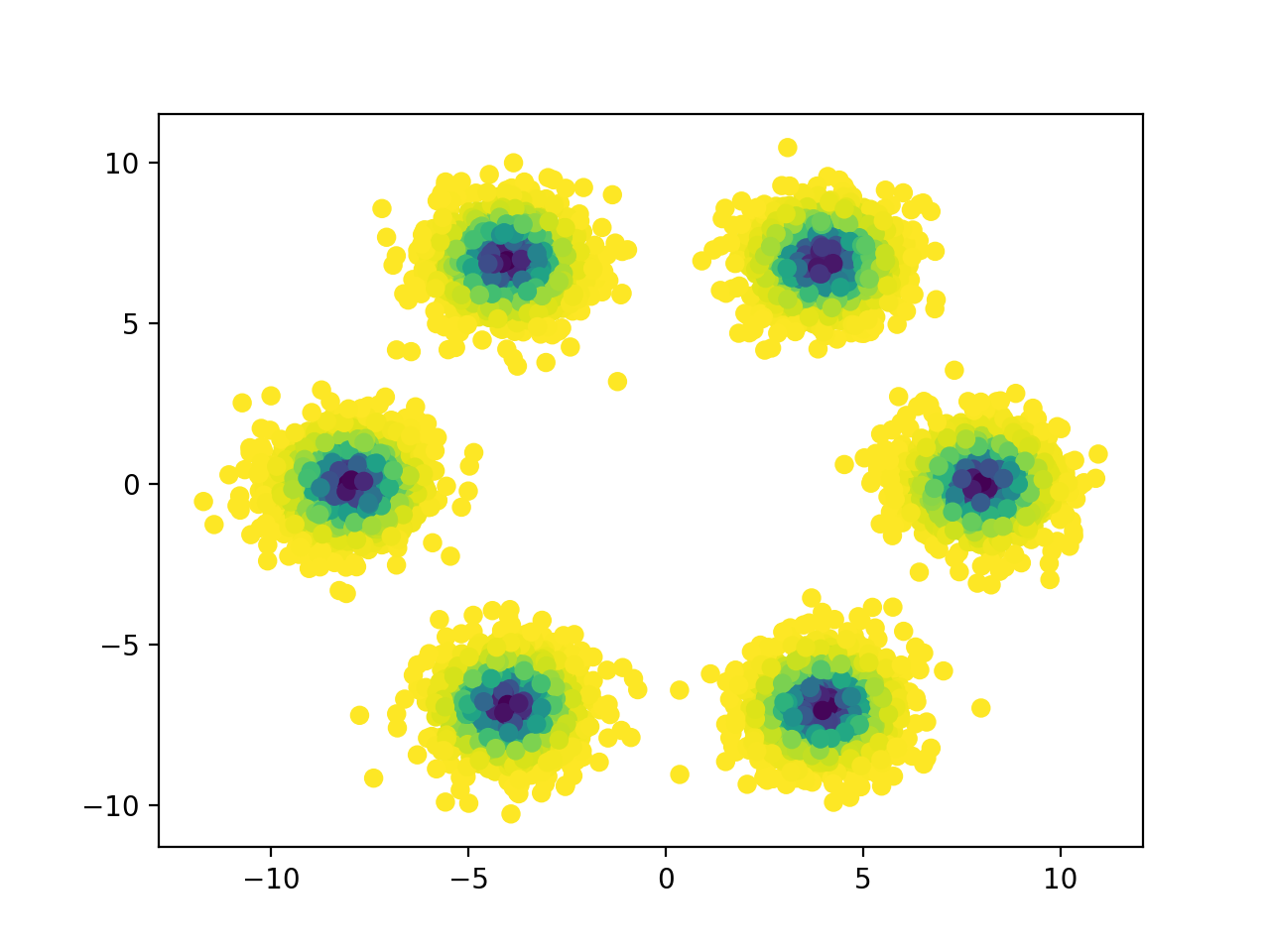} \\
(a) Initial $\bfx_i$   & (b) Latent data $\bfd_i$  & (c) Recovered $\widehat{\bfx}_i$
\end{tabular}
\caption{Same as Figure \ref{figdata2d} but with a new sample 
of $\bfx_i$. The recovery is done using the same potential
trained for Figure \ref{figdata2d}. \label{figdata2dVal}}
 \end{figure}

We present an example to show that our method can recover  a multi-modal distribution without the need to compute the partition function. 
We define the true potential $\phi_\mathrm{true}$ as the logarithm of the density
$p_G$ of a six-component Gaussian mixture on the
plane. More precisely, let
\[
p_G(\bfx) = \medmath{\frac{1}{6}\,\sum_{i=1}^6 }\,p_i(\bfx),
\]
where $p_i$ is the density of $N_2(\boldsymbol{\mu}_i,\bfI)$ for some mean vectors $\boldsymbol{\mu}_i$ 
(the centers of the clouds in Figure \ref{figdata2d}). Then, 
$$ \phi_\mathrm{true}(\bfx) = \log p_G(\bfx).$$
Panel (a) in Figure \ref{figdata2d} 
shows 600 samples $\bfx_i$ from this Gaussian mixture.
We use our method to construct a potential $\phi$ of the form \eqref{potential} based on the data $\bfx_i$ and compare it to the true potential. Unlike $\phi_\mathrm{true}$, the potential $\phi$
is a function of $\bfu$ and some estimated parameters $\bftheta$, so a direct comparison of these functions is not straightforward. We compare samples from the Gaussian mixture and their corresponding predictions based on the density defined by $\phi$.
The sampling from the latter distribution is done using recovery with the noisy points presented  in Algorithm~\ref{algForu}. The samples are shown in Figure \ref{figdata2d},
where the original $\bfx_i$ and the corresponding latent data 
$\bfd_i = \bfx_i + \bfepsilon_i$ are shown in panels (a) and (b), respectively. The color of the dots represents the value of $\phi_\mathrm{true}$ at these points. Panel
(c) shows the points 
$\widehat{\bfx}_i\equiv \bfQ_\ell\bfK \widehat{\bfu}(\bfx_i)$ whose color represents the value
of $\phi$. We see a good agreement between the predicted
values $\widehat{\bfx}_i$ and the initial values $\bfx_i$
of the sample.
For a  validation check, we generate a second sample of 1000 points
$\bfx_i$ and recreate panel (c) using the potential obtained for Figure \ref{figdata2d}. The results are shown in 
Figure \ref{figdata2dVal}. Again we see good agreement between initial and recovered data.

\edits{At this point it is interesting to compare the computational cost of our approach to the commonly used approach of maximum likelihood estimation (MLE) .
At their core, both methods are based on stochastic gradient descent and therefore behave in a similar way, asymptotically. Nonetheless, experimentally one can immediately observe that the constants can be vastly different.
But while the algorithms differ, they do have one common computational currency, the estimation of the gradient of the potential $\phi$. Ultimately, any method used to estimate $\phi$ can be assessed by the number of $\grad \phi$ computations. 
The basic iteration of MLE discussed in \cite{pml2Book}[Sec. 24.2.1] requires the computation of the score 
$ \grad_{\bftheta} \phi(\bfx_s, \bftheta)$ at new points $\bfx_s$,
where $\bfx_s$ are sampling points (not the original data points) of the potential $\phi$. To obtain $\bfx_s$,  one typically uses the Langevin dynamics
$$ \bfx_{s_{j+1}} = \bfx_{s_{j}} - {\frac {\varepsilon^2} 2}\grad \phi(\bfx_{s_{j}}) + \varepsilon \,\bfz_j,  $$
where the $\bfz_j$ are i.i.d. $N(\boldsymbol{0},\bfI)$ and $\varepsilon$ is a hyper parameter.
In the experiments done in \cite{pml2Book}, hundreds to thousands of iterations of the Langevin dynamics are used to estimate $\bfx_s$; thus, one evaluates the gradient of $\phi$ many times just to advance a single step in the SGD algorithm. In a sense, any SGD step of MLE contains within it another SGD step for the Langevin dynamics. This is in contrast to our approach where each evaluation of the gradient allows us to do one step.}

\subsection{Image Recovery}
\begin{figure}[t]
\begin{center}
\begin{tabular}{cccc}
\centering
\includegraphics[width=2.25cm]{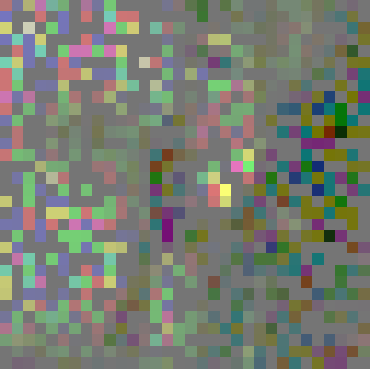} &
\includegraphics[width=2.3cm]{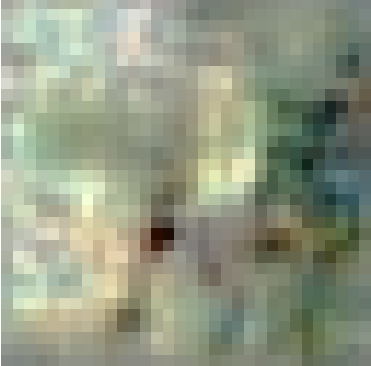} &
\includegraphics[width=2.20cm]{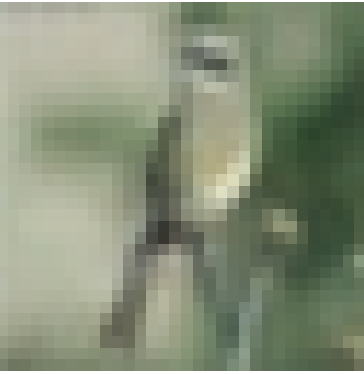} &
\includegraphics[width=2.25cm]{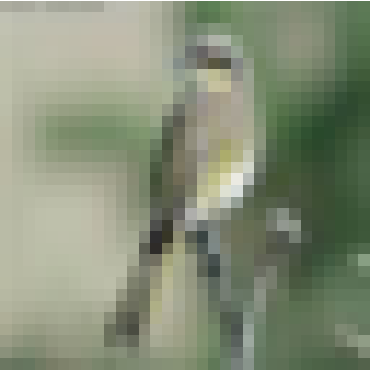}
\\
\includegraphics[width=2.25cm]{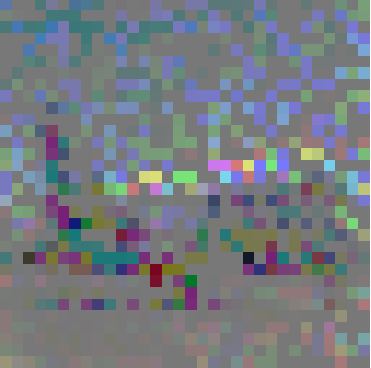} &
\includegraphics[width=2.25cm]{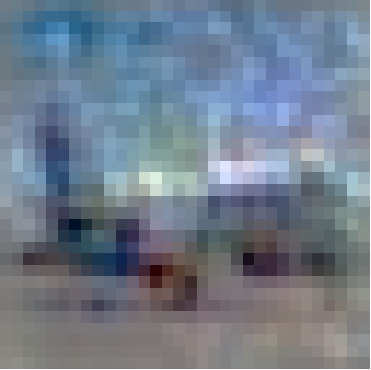} &
\includegraphics[width=2.25cm]{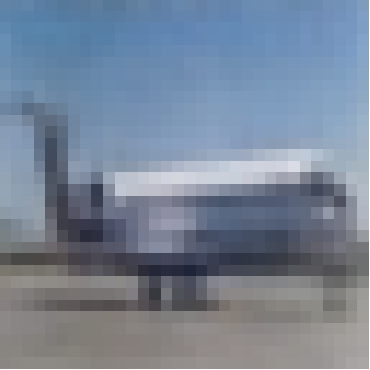} &
\includegraphics[width=2.25cm]{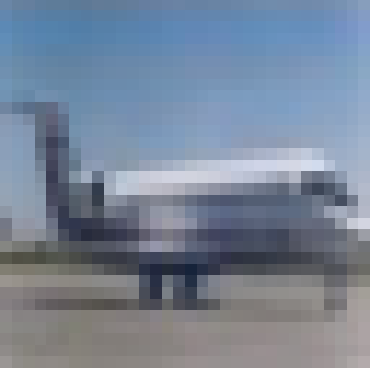}
\\
\includegraphics[width=2.25cm]{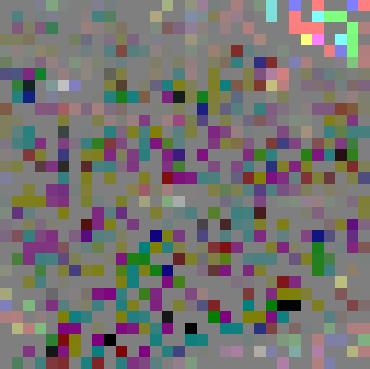} &
\includegraphics[width=2.25cm]{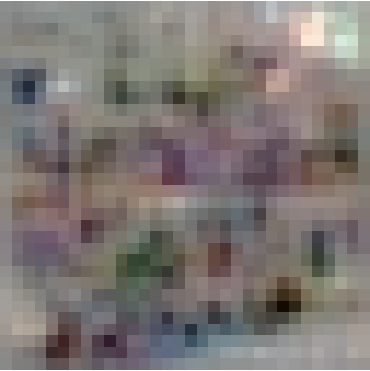} &
\includegraphics[width=2.25cm]{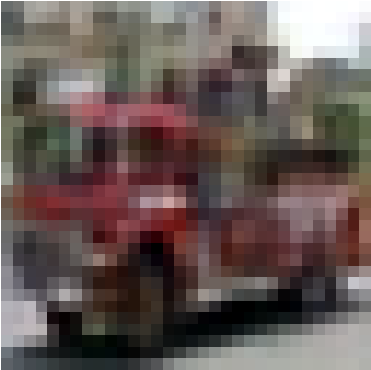} &
\includegraphics[width=2.25cm]{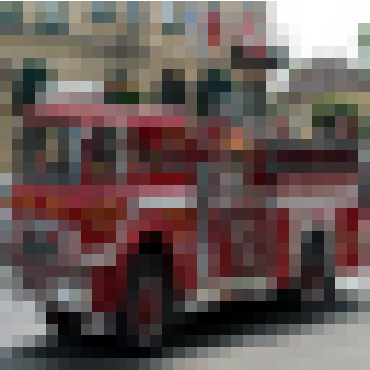}
\\
\includegraphics[width=2.25cm]{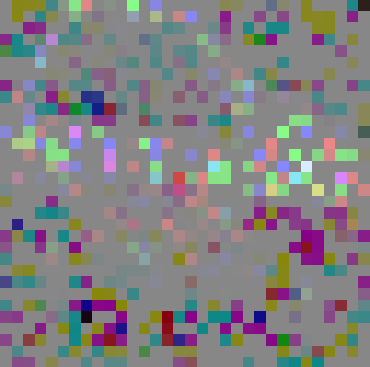} &
\includegraphics[width=2.25cm]{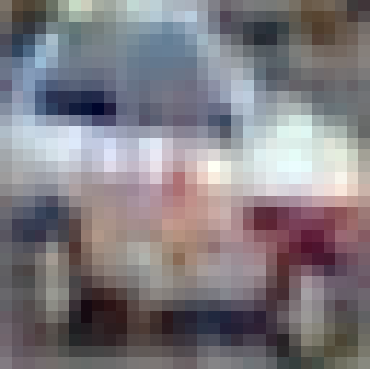} &
\includegraphics[width=2.25cm]{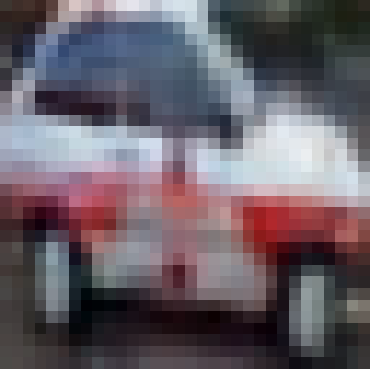} &
\includegraphics[width=2.25cm]{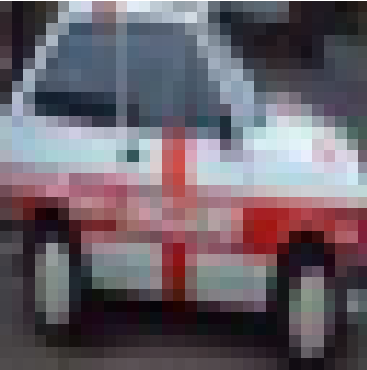}
\\
\includegraphics[width=2.25cm]{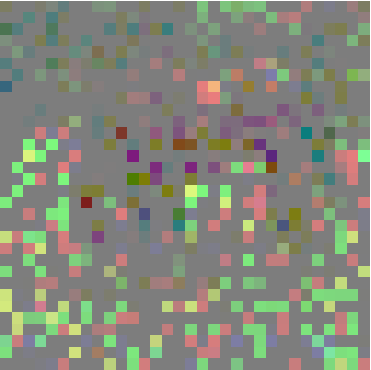} &
\includegraphics[width=2.25cm]{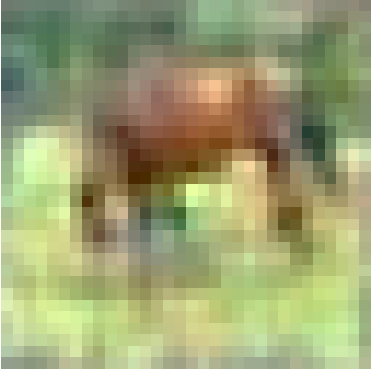} &
\includegraphics[width=2.25cm]{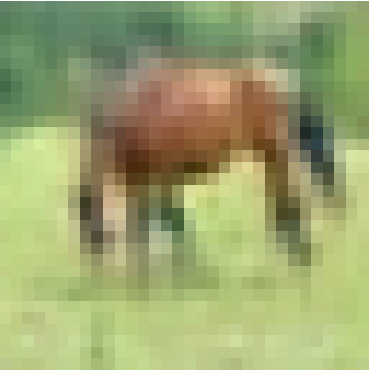} &
\includegraphics[width=2.25cm]{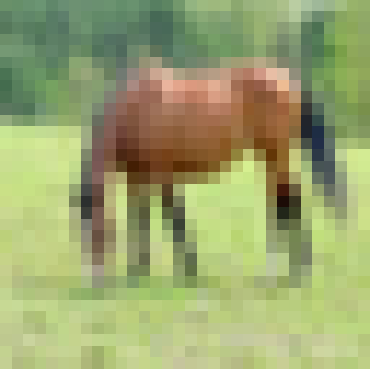}
\\
(a)  & (b)  & (c) & (d)  
\end{tabular}
\end{center}
\caption{Recovery of images from the CIFAR-10 test set. 
(a) Latent data $\bfd_i=\bfP_i\bfx_i+\bfepsilon_i$. 
(b) Data-fitted prediction $\bfK\bfu_0$. 
(c) Prediction $\widehat{\bfx}=\bfK\bfu_\ell$. (d) Original image $\bfx_i$.}
\label{fig:recovery_cifar10}
\end{figure}

\begin{figure}[t]
\begin{center}
\begin{tabular}{cccc}
\centering
\includegraphics[width=2.25cm]{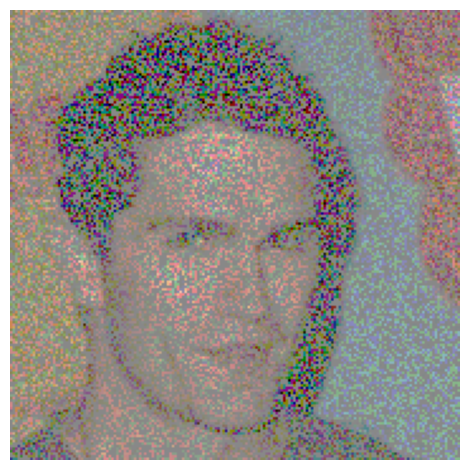} &
\includegraphics[width=2.25cm]{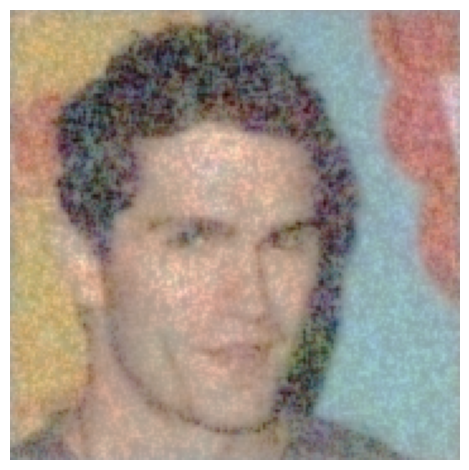} &
\includegraphics[width=2.25cm]{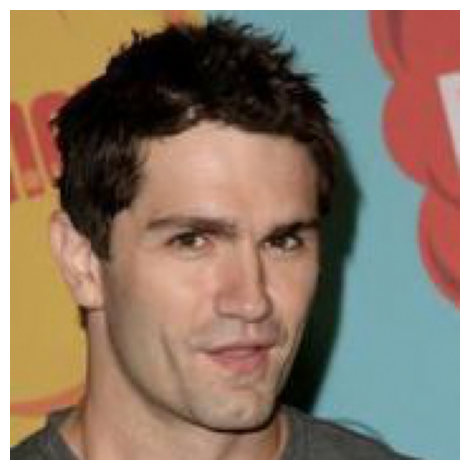} &
\includegraphics[width=2.25cm]{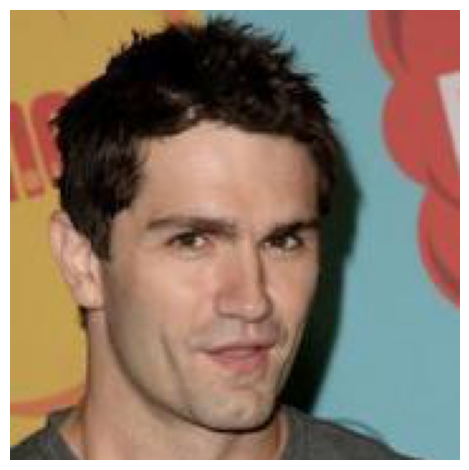}
\\
\includegraphics[width=2.25cm]{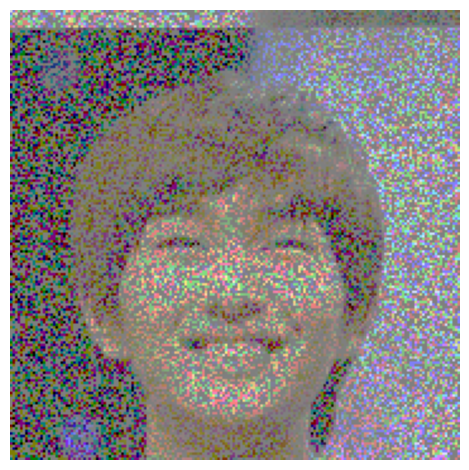} &
\includegraphics[width=2.25cm]{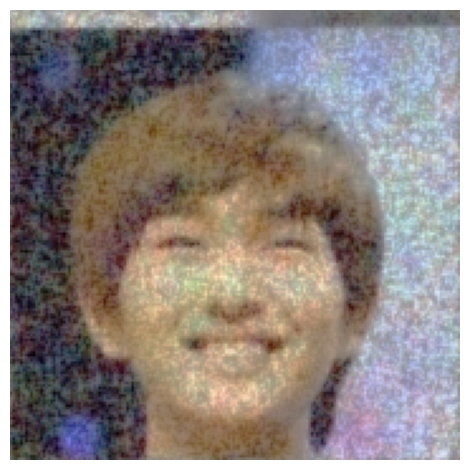} &
\includegraphics[width=2.25cm]{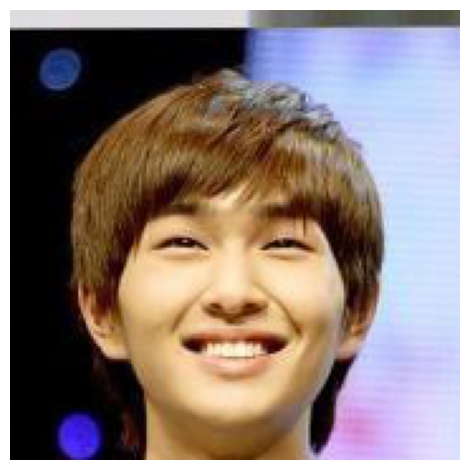} &
\includegraphics[width=2.25cm]{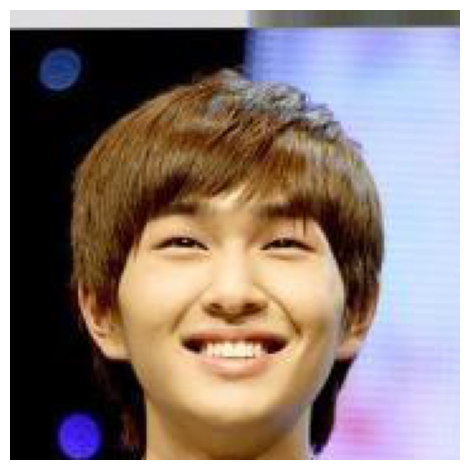}
\\
\includegraphics[width=2.25cm]{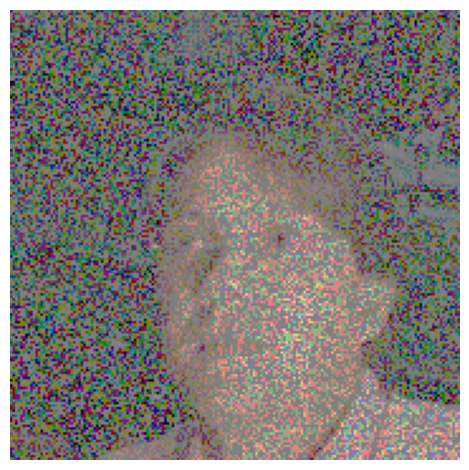} &
\includegraphics[width=2.25cm]{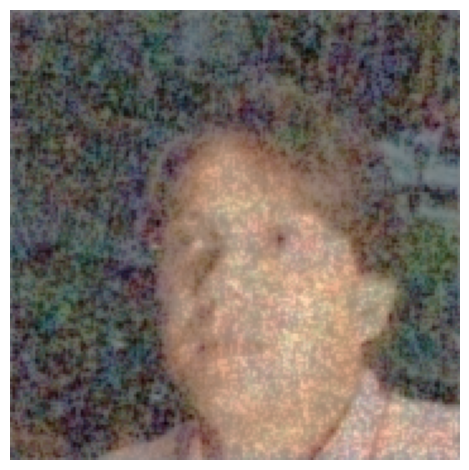} &
\includegraphics[width=2.25cm]{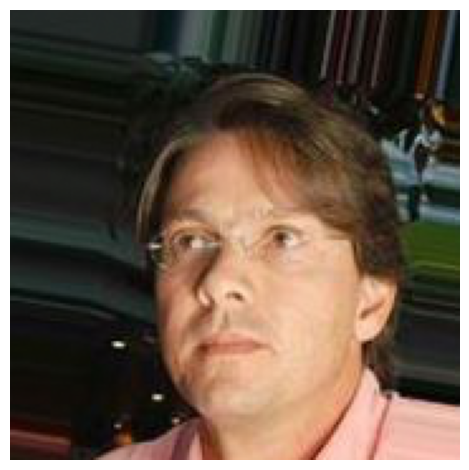} &
\includegraphics[width=2.25cm]{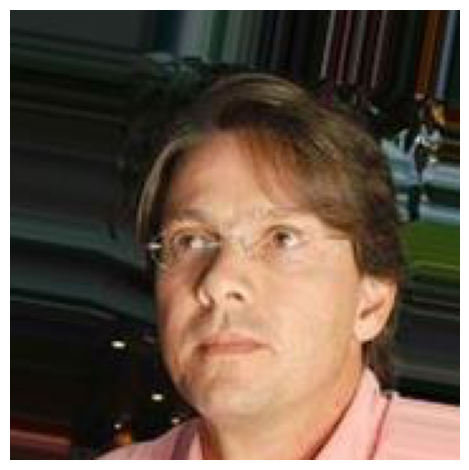}
\\
\includegraphics[width=2.25cm]{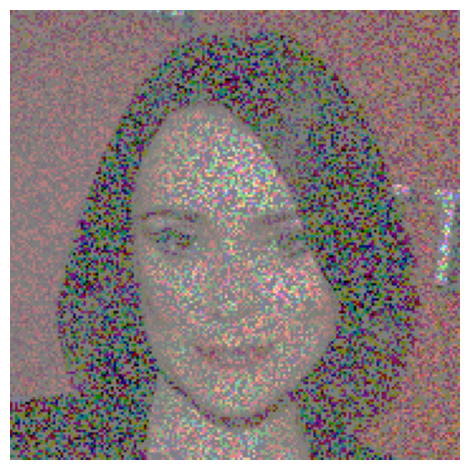} &
\includegraphics[width=2.25cm]{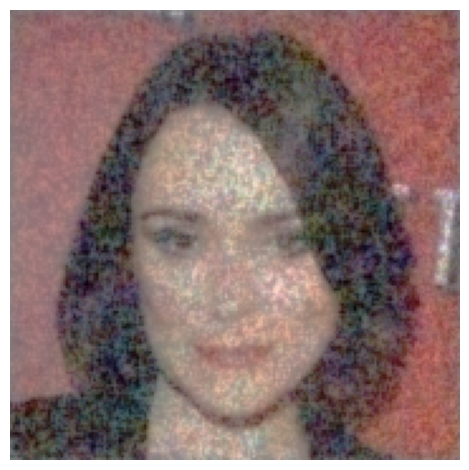} &
\includegraphics[width=2.25cm]{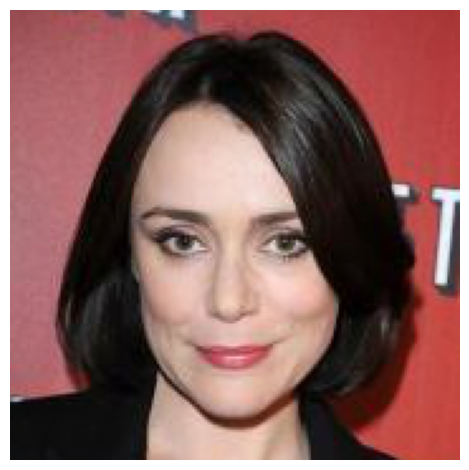} &
\includegraphics[width=2.25cm]{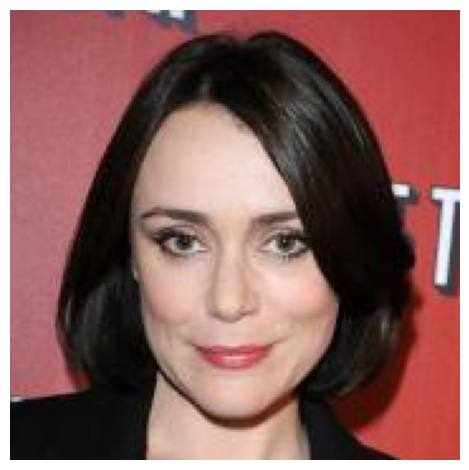}
\\
\includegraphics[width=2.25cm]{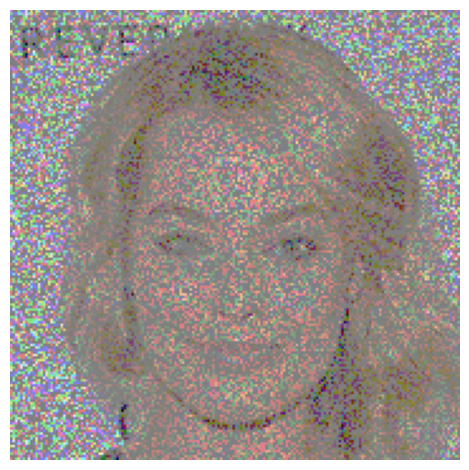} &
\includegraphics[width=2.25cm]{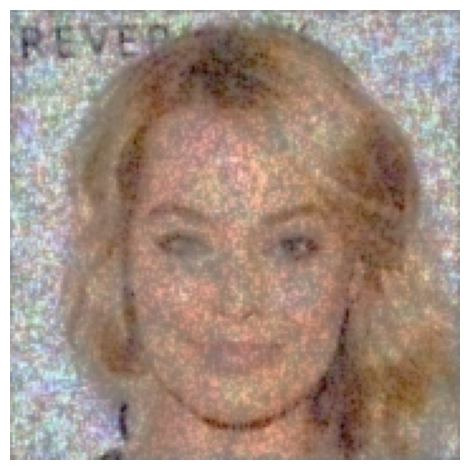} &
\includegraphics[width=2.25cm]{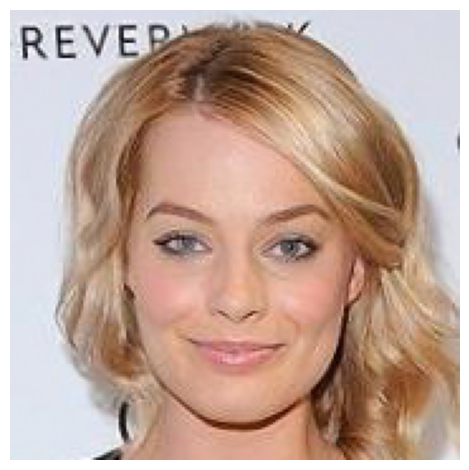} &
\includegraphics[width=2.25cm]{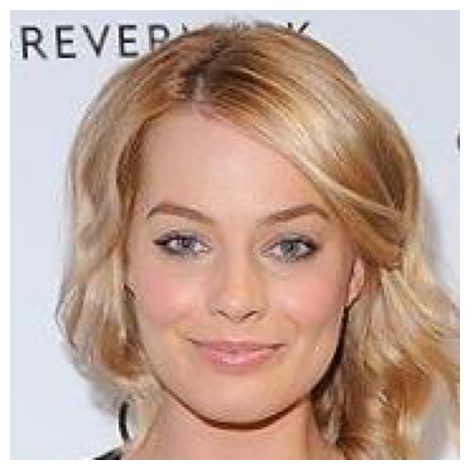}
\\
(a)  & (b)  & (c) & (d)  
\end{tabular}
\end{center}
\caption{Recovery of images from the CelebA test set. 
(a) Latent data $\bfd_i=\bfP_i\bfx_i+\bfepsilon_i$. 
(b) Data-fitted prediction $\bfK\bfu_0$. 
(c) Prediction $\widehat{\bfx}=\bfK\bfu_\ell$. (d) Original image $\bfx_i$. \label{fig:recovery_celebA}}
\end{figure}

We demonstrate the role played by the learned potential in the recovery of
images given partial information. More precisely, we use random matrices $\bfP$ that randomly select 30\% of the  pixels in the input image, and use the learned potential to recover the full image.
Figures~\ref{fig:recovery_cifar10} and \ref{fig:recovery_celebA} show that the prior defined by the learned potential recovers information lost during the random pixel selection. Columns (d) and (a) show, respectively, the original images $\bfx_i$ and latent data $\bfd_i=\bfP_i \bfx_i + \bfepsilon_i$. Column (b) shows the data-fitted prediction defined as $\bfK\bfu_0$ (see Algorithm \ref{algForu}) and the prediction $\widehat{\bfx}_i=\bfK\bfu_\ell$ (i.e., the application of the learned potential to the data-fitted prediction)
is shown in column (c).
The observed differences between columns (b) and (c) show the role of the learned potential; given a potentially partial recovery by data fitting, applying the learned potential adds missing details and improves the overall recovery quality.
The relative mean squared error of the recovered images are shown in the last column of Table~\ref{table:recovery}.

\begin{figure}[t]
\begin{center}
\begin{tabular}{ccccc}
\centering
\includegraphics[width=2.25cm]{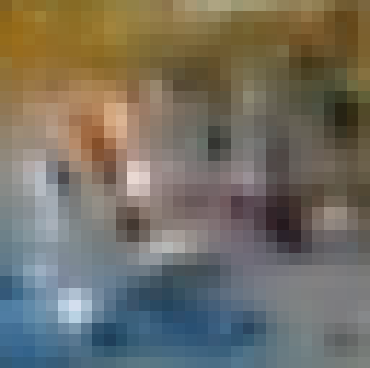}
&
\includegraphics[width=2.25cm]{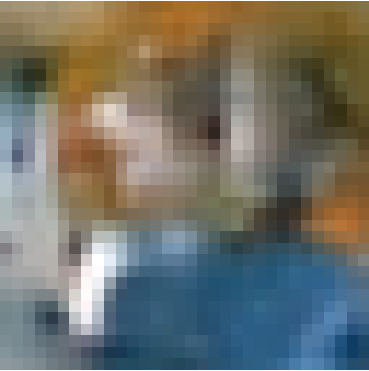} &
\includegraphics[width=2.25cm]{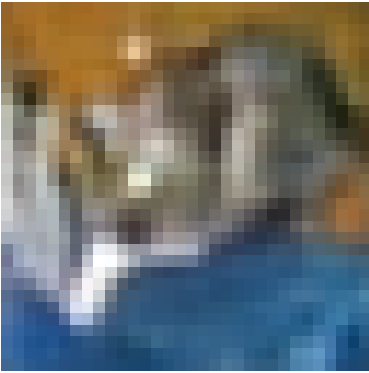} &
\includegraphics[width=2.25cm]{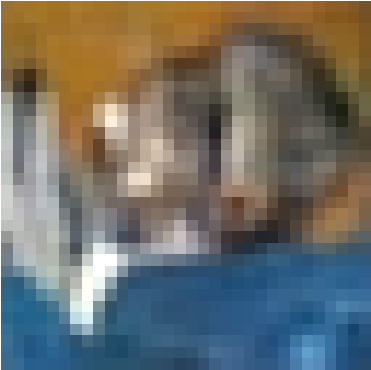} &
\includegraphics[width=2.25cm]{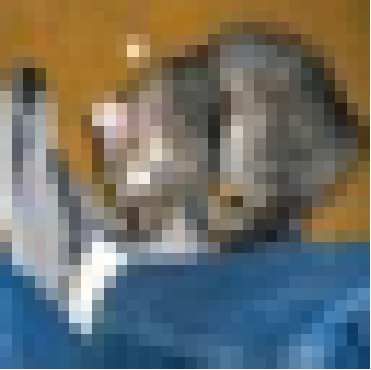}
\\

\includegraphics[width=2.25cm]{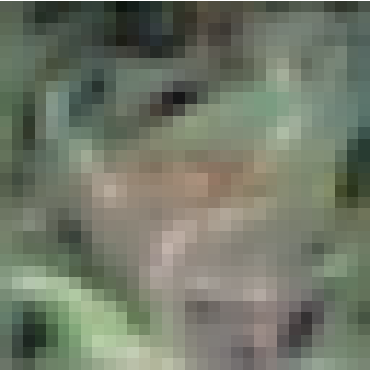}
&
\includegraphics[width=2.25cm]{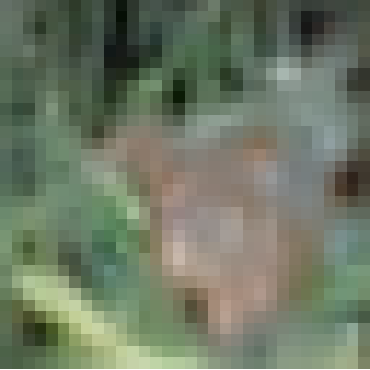} &
\includegraphics[width=2.25cm]{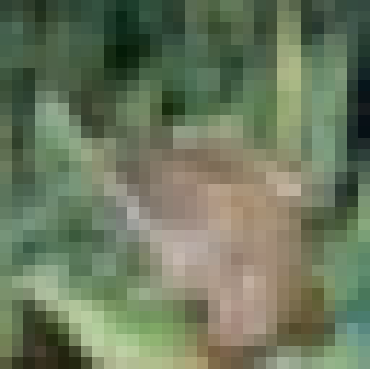} &
\includegraphics[width=2.25cm]{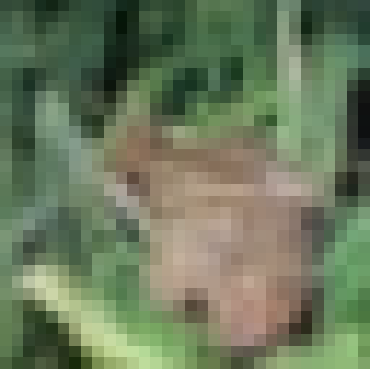} &
\includegraphics[width=2.25cm]{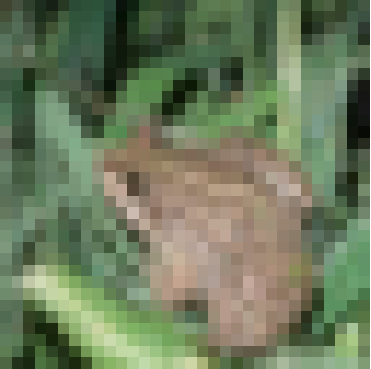}
\\

\includegraphics[width=2.25cm]{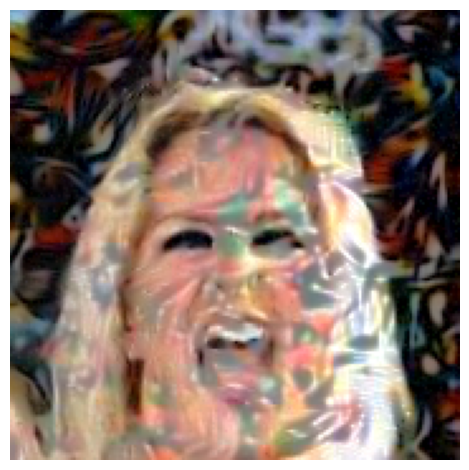}
&
\includegraphics[width=2.25cm]{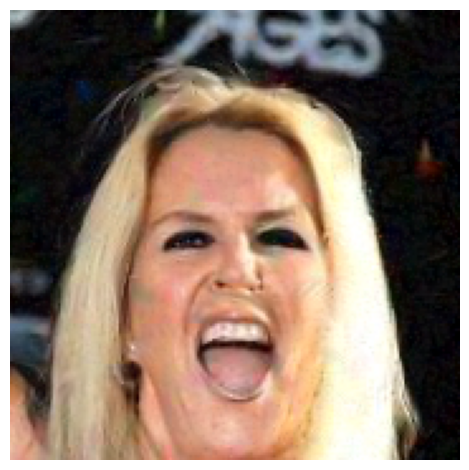} &
\includegraphics[width=2.25cm]{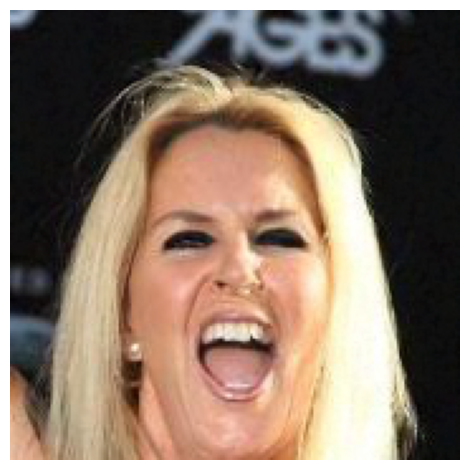} &
\includegraphics[width=2.25cm]{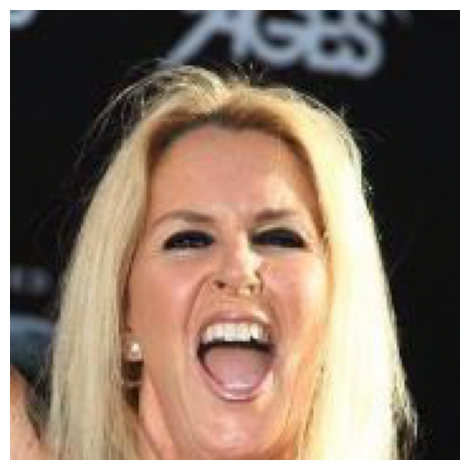} &
\includegraphics[width=2.25cm]{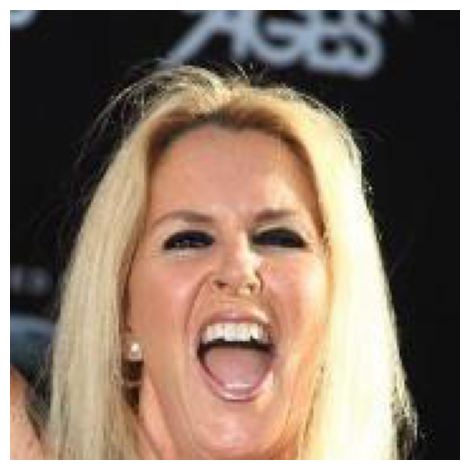}
\\

\includegraphics[width=2.25cm]{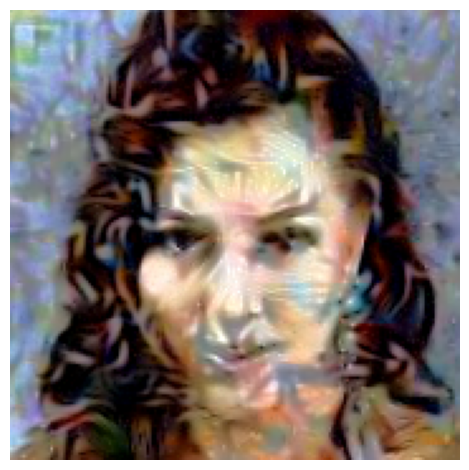}
&
\includegraphics[width=2.25cm]{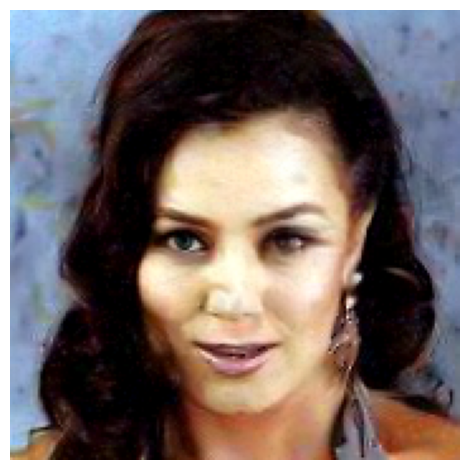} &
\includegraphics[width=2.25cm]{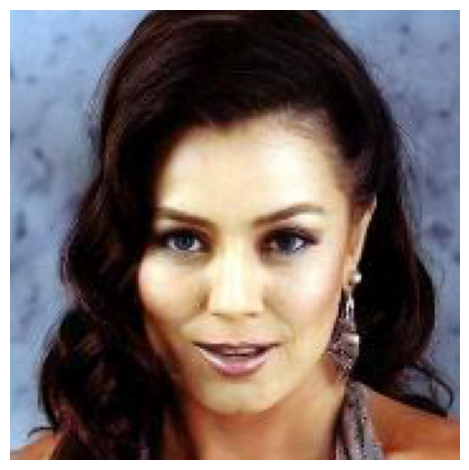} &
\includegraphics[width=2.25cm]{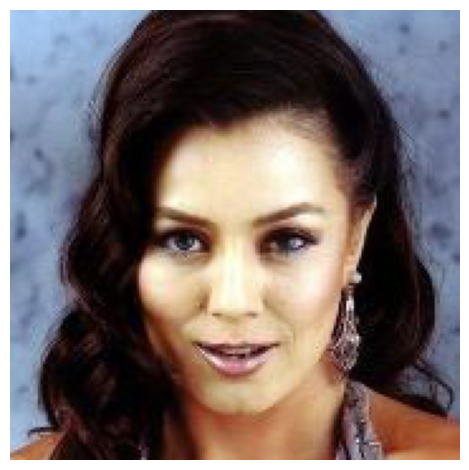} &
\includegraphics[width=2.25cm]{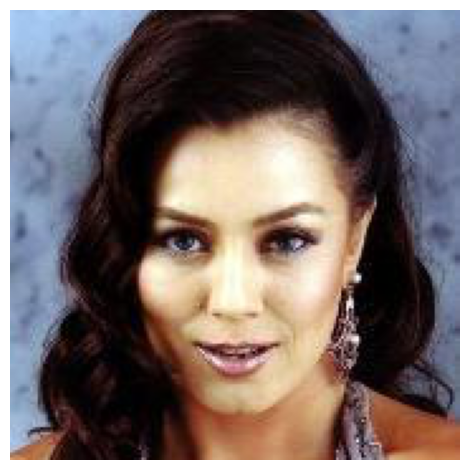}
\\

(a)  & (b)  & (c) & (d)  & (e)
\end{tabular}
\end{center}
\caption{Recovery of images from the CIFAR-10 and CelebA test sets using different percentages of observed pixels. (a) 5\%. 
(b) 10\%. (c) 20\%. (d) 30\%.  (e) Original image. \label{fig:recovery_celebA_various_percentages}}
\end{figure}

 To study the generalization properties of the learned potential, we remove different percentages of  pixels from the original images, but we use random matrices $\bfP$ that were  not used for the training. For every
image $\bfx$ in CIFAR-10 we use a random matrix 
$\bfP$ that selects a fraction of the pixels in $\bfx$ and add noise
to get a noisy partial image $\bfd = \bfP\bfx + \bfepsilon$, which in turn is used to obtain a reconstruction $\widehat{\bfx}$. We then compute the relative error 
$e=\|\widehat{\bfx} - \bfx\|^2/\|\bfx\|^2$. This is done 10 times for every image in CIFAR-10. We do the same with the  images in CelebA.
Table \ref{table:recovery}
summarizes the mean plus/minus standard deviation of all these
errors for different percentages of known pixels in each image. These means and standard deviations provide information about the size of the MSE and its variability across different images.
 \begin{table}[t]
  \caption{Relative recovery MSE.}
  \label{table:recovery}
  \begin{center}
  \begin{tabular}{|lcccc|}
  \hline
    Known &  \multirow{2}{*}{5\%} & \multirow{2}{*}{10\%} & \multirow{2}{*}{20\%} & \multirow{2}{*}{30\%}  \\
    pixels & & & &\\
    \hline
    CIFAR-10 & 3.51e-1 $\pm$ 3.47e-2  & 1.40e-1 $\pm$ 1.79e-2 & 4.41e-2 $\pm$ 5.29e-3  & 2.28e-2 $\pm$ 2.25e-3 \\
    CelebA & 6.74e-2 $\pm$ 6.02e-3 & 8.93e-3 $\pm$ 5.33e-4 & 1.21e-3 $\pm$ 1.61e-4& 8.07e-4 $\pm$ 4.29e-5\\
    \hline
  \end{tabular}
\end{center}
\end{table}
Another generalization check is shown in
Figure~\ref{fig:recovery_celebA_various_percentages}, 
which shows examples of recovered images from the test sets of CIFAR-10 and CelebA with varying percentages of available pixels.
We see that the potential generalizes to cases where significantly less data are available compared to the amount of
data in the training set.

\begin{figure}[t]
\begin{center}
\begin{tabular}{cccc}
\centering
\includegraphics[width=0.8\columnwidth]{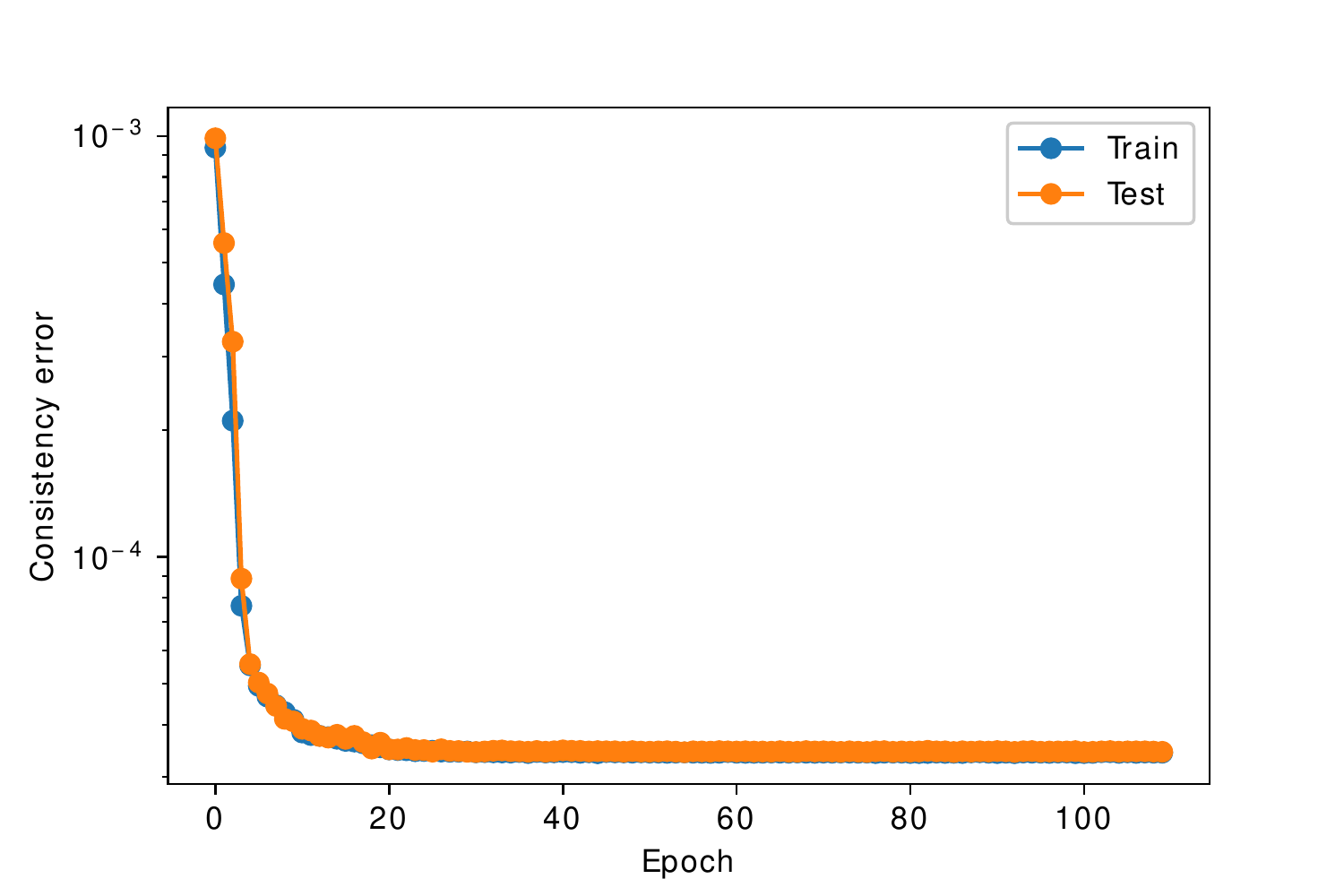} & 
\end{tabular}
\end{center}
\caption{Consistency error $R_c$ for CIFAR-10 as a function of the number of epochs.\label{fig:cifar10_consistecyloss}}
\end{figure}

An important feature of our network is its ability to approximate the 
boundary value problem \eqref{dmapu2} with an initial value problem.
Figure~\ref{fig:cifar10_consistecyloss} shows the error, as measured by $R_c$, in replacing the boundary value problem with the initial value problem. The training and validation errors are very small (machine precision) and it is learned after very few epochs.

\section{Conclusions}
\label{sec5}
Recovery of a potential is an important problem in computational statistical learning that is typically solved using maximum likelihood. 
We have seen that computing these estimates for general potentials
and high-dimensional problems is difficult because of the cost
of computing or approximating the partition function or its gradient. 
If we approximate the partition function as explained in Section \ref{sec1}, then it follows from 
\eqref{gradminlog} and \eqref{eq:ex-gradest} that for $\bftheta$ to be an extremum of the likelihood we need a sample $\widetilde{\bfx}_j$ with density $p_\bftheta$ to match two sample averages:
\[
\medmath{\frac{1}{N}\,\sum_{j=1}^N }\,\grad_{\bftheta} \phi(\widetilde{\bfx}_j,\bftheta)\approx 
{\mathbb E}\,\grad_{\bftheta} \phi(\bfx, \bftheta)        =
\medmath{\frac{1}{n} \,
\sum_{i=1}^n }\,\grad_{\bftheta} \phi(\bfx_i,\bftheta).
\]
The difficulty with this technique  is the computation of the samples $\widetilde{\bfx}_j$ for different values of $\bftheta$ which requires long iterative procedures.
 We take a different approach where we try to match the data $\bfx_i$
 themselves using predictions based on latent data.
Our methodology, that we call MR-MAP, can be used  in  the parametric case
where the potential is known up to parameters, or in the case
where the density is unknown and the form of the potential
has to be determined. In the parametric case, an estimate of the parameters $\bftheta$ is defined by finding the parameters that minimize an estimate of the MSE of a MAP
of the posterior distribution of $\bfx_i$ given some latent
data, which we  define as noisy projections of $\bfx_i$.
When the form of the potential is unknown, we use a least-action potential whose gradient with respect to the flow is a multilayer neural network with parameters estimated as in the parametric case. To solve the optimization problem, which
includes the neural networks and the minimization of the MSE
estimate, we use the simple structure of a so called hyperbolic neural network 
that uses learned initial conditions to propagate  forward and obtain a solution.

The performance of our methodology was illustrated 
using a small synthetic 
problem as well as data sets of natural images (CIFAR10)
and human faces (Celeb-A).
 We showed that the network yields good recovery of the potential and of the images given partial data.
Even though the potential was trained with only 30\% of the pixels in the image, we were able to obtain reasonable recovery for problems 
where only 5\% of the pixels are present. 

We believe our approach has many promising  applications. In particular, we are in the process of applying it to regularization of inverse problems  as well as to problems in biology that require the estimation of a potential function given observed  data.

 \appendix

 \section{Regarding consistency}\label{sec:consistency}
\edits{
It is well known that maximum likelihood estimators (MLEs) may not be consistent; for example there are simple Gaussian problems where there is a consistent moment estimator but the MLE is inconsistent
(see, e.g., \cite{neyman-scottMLE} and \cite{lecam-book}[Ch.17.8]). But MLEs can be shown to be strongly consistent under
somewhat restrictive regularity conditions.
 Strong consistency of $\widehat{\bftheta}_n$ based on i.i.d. samples  $\bfx_1,\ldots,\bfx_n$ in $\mathbb{R}^p$ can also be proved under analogous regularity conditions by adapting Wald's
 classic proof of MLE consistency  \cite{waldMLE}  and a uniform  strong law.
 We just have to change a likelihood ratio to a MSE difference, and reformulate the condition of identifiability through the likelihood to identifiability through a Bayes risk. For completeness, we include a proof here.}

\edits{
 Let $\widehat{\bfx}(\bftheta)$ be a MAP corresponding to a prior density $p_\bftheta$ as defined in Section \ref{sec2} and set
 $M(\bfx,\bftheta) = \mathbb{E}\,(\,\|\widehat{\bfx}(\bftheta) -\bfx\,\|^2\mid \bfx\,)$ 
 and $B(\bftheta)  = \mathbb{E}_{\bftheta_0} M(\bfx,\bftheta)$
 for $\bftheta\in \Theta\subset \mathbb{R}^d$.
As usual, the subscript in the expectation means that $\bfx$ has 
density $p_{\bftheta_0}$, where $\bftheta_0$ is the true parameter.
The following assumptions are analogous to those used in proofs of  MLE consistency (see, e.g., \cite{ferguson-book}[Ch.17]):}
\edits{
\begin{itemize}
\item[(i)] $M$ is Borel measurable in $(\bfx,\bftheta)$.
\item[(ii)] $\Theta$ is a compact set and $B$ is bounded on $\Theta$.
\item[(iii)] For each $\bfx$, $\|\widehat{\bfx}(\bftheta)-\bfx\|^2$ is upper semicontinuous in $\bftheta$.
 \item[(iv)]  There is a function $K$ such that $M(\bfx,\bftheta)\leq K(\bfx)$  for all $\bfx\in \mathbb{R}^p$ and $\bftheta\in \Theta$,
  and $\mathbb{E}_{\bftheta} K(\bfx)<\infty$ for all $\bftheta\in \Theta$.
 \item[(v)] $B$ has a unique global minimum at $\bftheta_0$ (i.e., $\bftheta_0$ is `Bayes-risk identifiable' ).
\item[(vi)] For all $\bftheta\in \Theta$ and all small enough $\varepsilon>0$, the function 
  $\bfx\to \sup_{\|\bftheta'-\bftheta\|<\varepsilon}M(\bfx,\bftheta')$ is Borel measurable.
\end{itemize}
}

\vspace{.2cm}

\edits{
 Set  $D(\bfx,\bftheta) = M(\bfx,\bftheta_0) - M(\bfx,\bftheta)$ and $d(\bftheta) = \mathbb{E}_{\bftheta_0} D(\bfx,\bftheta)$.
 Clearly, $D(\bfx,\bftheta_0)=0$ for all $\bfx$, and thus by the definition of $\widehat{\bftheta}_n$  we have 
  \begin{equation}\label{eq:nonnD}
 \medmath{\frac{1}{n}\,\sum_{k=1}^n}\, D(\bfx_k,\widehat{\bftheta}_n) = 
 \sup_{\bftheta\in \Theta}\,\medmath{ \frac{1}{n}\,\sum_{k=1}^n}\, D(\bfx_k,\bftheta)\geq 0.
 \end{equation}
 For $\varepsilon>0$ define 
 \(
 S_\varepsilon = \{\theta\in \Theta: \|\bftheta-\bftheta_0\|\geq \varepsilon\}.
 \)
 It follows from \eqref{eq:nonnD} that for any $\delta<0$,
 \begin{eqnarray*}
 \mathbb{P}_{\bftheta_0}\left(\,\limsup_n\,\sup_{\bftheta\in S_\varepsilon} \medmath{\frac{1}{n}\,\sum_{k=1}^n}\, D(\bfx_k,\bftheta)
 < \delta \,\right) &\leq & 
 \mathbb{P}_{\bftheta_0}
 \left(\,
 \medmath{\bigcup_{N\geq 1}\,\bigcap_{n\geq N}}\,
 \left\{\,
\sup_{\bftheta\in S_\varepsilon} \medmath{\frac{1}{n}\,\sum_{k=1}^n}\, 
 D(\bfx_k,\bftheta) \leq \delta
 \,\right\}
 \,\right)\\
 &\leq & \mathbb{P}_{\bftheta_0}
 \left(\,
 \medmath{\bigcup_{N\geq 1}\,\bigcap_{n\geq N}}\,
 \left\{\,
 \|\widehat{\bftheta}_n-\bftheta_0\|<\varepsilon
 \,\right\}
 \,\right).
 \end{eqnarray*}
 To prove $\widehat{\bftheta}_n\,\arras\,\bftheta_0$, it is therefore sufficient  to show that the probability on the left hand-side is one for some
 $\delta<0$.
 To show this, note that from assumptions (iii), (iv) and Fatou's lemma (for the limsup) it follows that $D(\bfx,\bftheta)$ is upper semicontinuous in $\bftheta$ for each $\bfx$. Similarly, $B$ is upper
 semicontinuous on $\Theta$. The assumed regularity conditions also imply the following uniform strong law 
 (see, e.g., \cite{ferguson-book}[Ch.16]): For any compact set $S\subset \mathbb{R}^d$,
 \begin{equation}\label{eq:unslln}
 \mathbb{P}_{\bftheta_0}\left(\,\limsup_n\,\sup_{\bftheta\in S} \medmath{\frac{1}{n}\,\sum_{k=1}^n}\, D(\bfx_k,\bftheta)
 \leq \sup_{\bftheta\in S} B(\bftheta)\,\right) = 1.
 \end{equation}
 Since $\Theta$ is compact, so is  $S_\varepsilon$ and thus, by upper semicontinuity, the maximum of $B$ on $S_\varepsilon$ is achieved in $S_\varepsilon$. Also, by assumption (v),
 \(
 B_\varepsilon^* = \sup_{\bftheta\in S_\varepsilon} B(\bftheta) < 0.
 \)
 Fix $\delta\in (B_\varepsilon^*, 0)$. From \eqref{eq:unslln} we get
\[
 1=\mathbb{P}_{\bftheta_0}\left(\,\limsup_n\,\sup_{\bftheta\in S_\varepsilon} \medmath{\frac{1}{n}\,\sum_{k=1}^n}\, D(\bfx_k,\bftheta)
 \leq B_\varepsilon^*\,\right)\leq
 \mathbb{P}_{\bftheta_0}\left(\,\limsup_n\,\sup_{\bftheta\in S_\varepsilon} \medmath{\frac{1}{n}\,\sum_{k=1}^n}\, D(\bfx_k,\bftheta)
 < \delta\,\right), 
\]
which proves \,$\widehat{\bftheta}_n\,\,\arras\,\,\bftheta_0$.
}

\bibliographystyle{siamplain}
\bibliography{biblio}

\end{document}